\documentclass{article}

\usepackage{arxiv}
\usepackage[utf8]{inputenc} 
\usepackage[T1]{fontenc}    
\usepackage{hyperref}       
\usepackage{url}            
\usepackage{booktabs}       
\usepackage{amsfonts}       
\usepackage{nicefrac}       
\usepackage{microtype}      
\usepackage{lipsum}
\usepackage{graphicx}
\usepackage{subfiles}
\graphicspath{ {./images/} }
\usepackage{authblk}
\usepackage{xcolor}
\usepackage{amsmath}
\usepackage{amssymb}
\usepackage{flafter}             
\usepackage[section]{placeins}   
\usepackage{dblfloatfix}        
\usepackage{tikz}
\usetikzlibrary{positioning,arrows.meta,fit,calc,shapes.geometric}
\usepackage{mathrsfs}
\usepackage{float}
\usepackage{subcaption}
\usepackage{listings}
\usepackage{xcolor}

\lstdefinestyle{promptstyle}{
  basicstyle=\ttfamily\footnotesize,
  breaklines=true,
  breakatwhitespace=true,
  columns=fullflexible,
  frame=single,
  framerule=0.4pt,
  rulecolor=\color{black!25},
  backgroundcolor=\color{black!2},
  xleftmargin=0.8em,
  xrightmargin=0.4em,
  aboveskip=0.8em,
  belowskip=0.8em,
  showstringspaces=false,
  tabsize=2
}

\lstdefinestyle{jsonstyle}{
  style=promptstyle,
  basicstyle=\ttfamily\scriptsize,
  breakatwhitespace=false, 
  keepspaces=true,         
  showstringspaces=false
}

\newcommand{\up}[1]{\textcolor{green!55!black}{$\uparrow\,#1$}}
\newcommand{\down}[1]{\textcolor{red!70!black}{$\downarrow\,#1$}}

\definecolor{myBlue}{RGB}{218, 232, 252}
\definecolor{myOrange}{RGB}{255, 230, 204}
\definecolor{myGreen}{RGB}{234, 241, 221} 
\definecolor{myGray}{RGB}{245, 245, 245}
\definecolor{myDarkGray}{RGB}{210, 210, 210}
\newcommand{\wenxiuedit}[1]{\textcolor{black}{#1}}

\title{AIDABench: AI Data Analytics  Benchmark}

\author[1*]{Yibo Yang}
\author[1*]{Fei Lei}
\author[1]{Yixuan Sun}
\author[1]{Yantao Zeng}
\author[1]{Chengguang Lv}
\author[1]{Jiancao Hong}
\author[1]{Jiaojiao Tian}
\author[1]{Tianyu Qiu}
\author[1]{Xin Wang}
\author[1]{Yanbing Chen}
\author[1]{Yanjie Li}
\author[1]{Zheng Pan}
\author[1]{Xiaochen Zhou}
\author[1]{Guanzhou Chen}
\author[1]{Haoran Lv}
\author[1]{Yuning Xu}
\author[]{Yue Ou}
\author[]{Haodong Liu}
\author[]{Shiqi He}
\author[]{Anya Jia}
\author[]{Yulei Xin}
\author[]{Huan Wu}
\author[]{Liang Liu}
\author[2]{Jiaye Ge}
\author[2]{Jianxin Dong}
\author[1,3]{Dahua Lin}
\author[1*]{Wenxiu Sun}
\affil[1]{SenseTime Research}
\affil[2]{Shanghai AI Laboratory}
\affil[3]{MMLab, The Chinese University of Hong Kong}
\affil[*]{\texttt{\{yangyibo1, leifei1, sunwx\}@sensetime.com}}

\begin{document}
\maketitle
\begin{abstract}
As AI-driven document understanding and processing tools become increasingly prevalent in real-world applications, the need for rigorous evaluation standards has grown increasingly urgent. Existing benchmarks and evaluations often focus on isolated capabilities or simplified scenarios, failing to capture the end-to-end task effectiveness required in real-world settings. To address this gap, we introduce AIDABench, a comprehensive benchmark for evaluating AI systems on complex Data Analytics tasks in an end-to-end manner. AIDABench encompasses $600+$ diverse document analytical tasks across three core capability dimensions: question answering, data visualization,  and file generation. These tasks are grounded in realistic scenarios involving heterogeneous data types—including spreadsheets, databases, financial reports, and operational records—and reflect the analytical demands encountered across diverse industries and job functions. Notably, the tasks in AIDABench are sufficiently challenging that even human experts require $1$-$2$ hours per question when assisted by AI tools, underscoring the benchmark's difficulty and real-world complexity.
We evaluate 11 state-of-the-art models, \wenxiuedit{spanning both proprietary (e.g., claude-sonnet-4-5, gemini-3-pro-preview) and open-source (e.g., qwen3-max-2026-01-23-thinking) families} on AIDABench. Our results reveal that complex, real-world data analytics tasks remain a significant challenge for current AI systems, with the best-performing model achieving only \wenxiuedit{$59.43$ pass@1}. We provide a detailed analysis of failure modes across each capability dimension and identify key challenges for future research. AIDABench offers a principled reference for enterprise procurement, tool selection, and model optimization, and is publicly available at \url{https://github.com/MichaelYang-lyx/AIDABench}.

\end{abstract}


\section{Introduction }
\label{sec:intro}

\begin{figure}[!t]
    \centering
    \includegraphics[width=\textwidth, keepaspectratio]{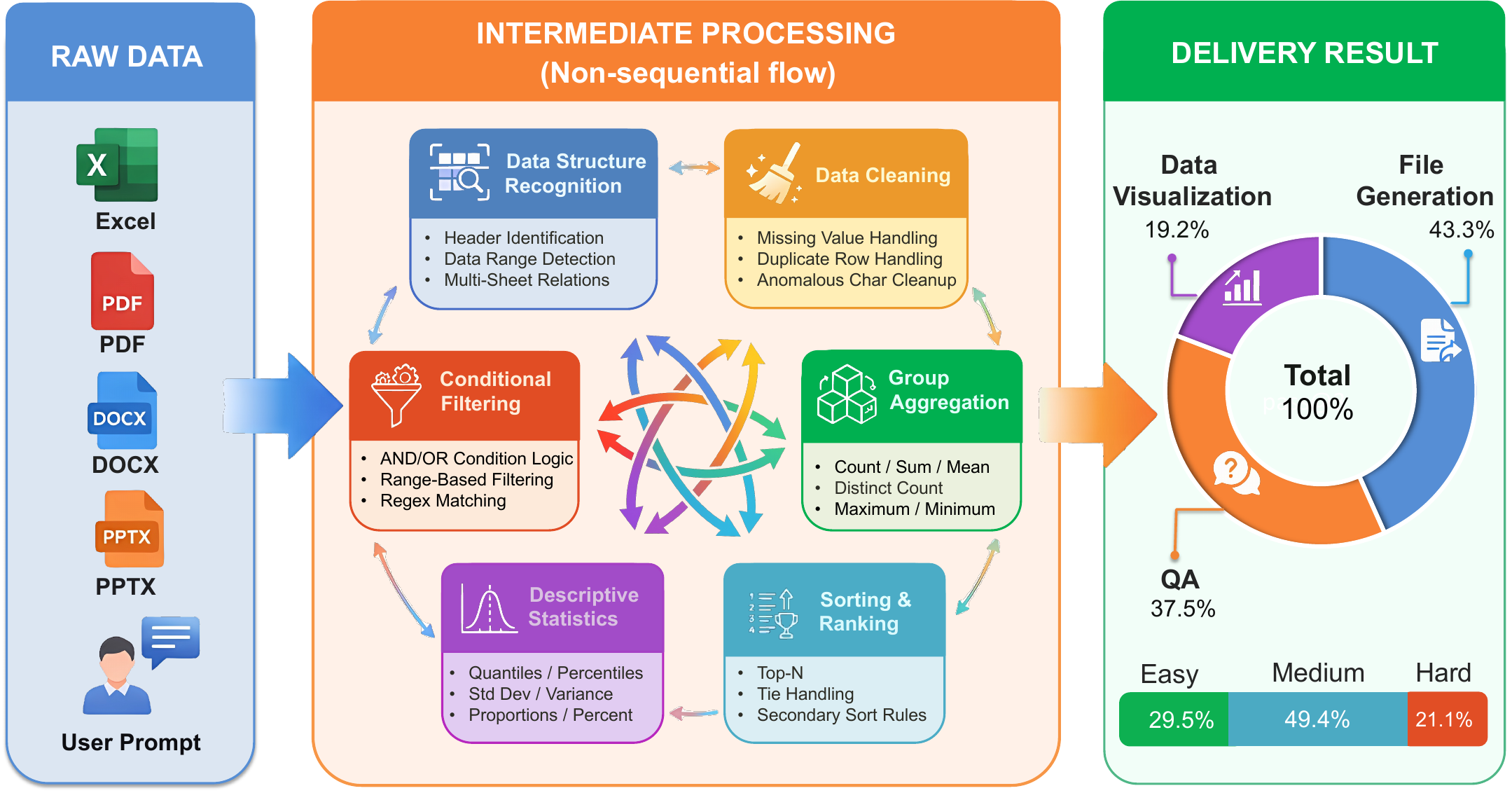}
    \caption{Overview of the AIDABench evaluation framework. The workflow illustrates the pipeline from multi-format Raw Data ingestion, through Intermediate Processing (encompassing the capability dimensions of data editing/transformation and numerical/statistical reasoning in a non-sequential flow), to the final Delivery Result (corresponding to the QA, data visualization, and file generation dimensions).}
    \label{fig:overview}
\end{figure}

As AI-driven document analysis agents evolve from basic text processing to autonomous, end-to-end handling of complex multi-step data workflows, the need for comprehensive evaluation standards has become increasingly pressing.
Such workflows handle a wide spectrum of document types commonly encountered in professional settings, ranging from structured data sources such as spreadsheets, CSV files, and database records, to semi-structured business documents including sales reports, financial statements, invoices, attendance logs, and project tracking records. By interpreting natural-language instructions, these tools can autonomously execute core operations—such as data querying, statistical computation, data cleaning, reformatting, and visualization—thereby streamlining critical tasks like report generation, data preparation, and decision support.

In recent years, AI-driven document analysis agents have proliferated rapidly. However, many existing evaluations emphasize performance on isolated dimensions or narrowly defined scenarios under simplified settings, providing only partial evidence that may fail to capture real-world robustness and end-to-end task effectiveness. Such single-dimensional assessments are insufficient to faithfully characterize practical tool competence, limiting their usefulness for model selection, industry adoption, and procurement decisions. To address these limitations, we introduce \textbf{AIDABench}, a comprehensive benchmarking framework designed to provide enterprises, research and engineering teams, and frontline practitioners with a scientifically grounded and holistic evaluation standard for AI-driven document analysis tools. AIDABench is built around realistic scenarios featuring heterogeneous data types and complex analytical requirements, covering the end-to-end document processing pipeline while remaining adaptable to diverse industries and job functions.

AIDABench serves as a rigorous benchmarking suite for evaluating LLM-driven document analysis tools, with an emphasis on core data-processing capabilities required in real-world scenarios. In terms of difficulty distribution, AIDABench consists of $29.5\%$ easy tasks, $49.4\%$ medium tasks, and $21.1\%$ hard tasks that often require $13$ or more reasoning steps to complete. The evaluation protocol comprehensively covers high-frequency data processing tasks and is structured around three primary capability dimensions, as shown in Fig.~\ref{fig:overview}.
\textbf{Question Answering.} This dimension evaluates a broad range of analytical operations commonly required in document analysis, including summation, mean computation, proportion and share analysis, ranking, extrema extraction, dispersion measures, and trend analysis, thereby addressing essential numerical reasoning demands in real-world data processing scenarios.
\textbf{File Generation.} This dimension assesses high-utility data wrangling operations such as filtering, format normalization, content correction, batch replacement, cross-sheet linkage and coordinated updates, deduplication, and missing-value imputation, reflecting frequent procedures in practical data preparation and cleaning workflows.
\textbf{Data Visualization.} This dimension measures the ability to generate and adapt multiple visualization forms—including bar charts, line charts, pie charts, and tabular presentations—along with style customization, to meet the requirements of data communication and presentation in reporting and briefing contexts.

The key contributions of our work are summarized as follows:
\begin{itemize}
    \item We introduce AIDABench, a comprehensive benchmarking framework 
    for evaluating LLM-driven document analysis agents across the 
    end-to-end data processing pipeline, featuring heterogeneous 
    document types and realistic analytical requirements.

    \item We design a multi-dimensional evaluation protocol structured 
    around three primary capability dimensions—Question Answering, 
    Data Visualization, and File Generation—with a balanced difficulty 
    distribution (29.5\% easy, 49.4\% medium, and 21.1\% hard) 
    to ensure thorough assessment of model competence.

    \item We develop specialized evaluators for three task categories that closely reflect human judgment, and conduct extensive experiments on 11 state-of-the-art LLMs spanning diverse model families, yielding fine-grained analyses and actionable insights for enterprise procurement, tool selection, and model-level optimization.
\end{itemize}

\section{Related Works }
\label{sec:related}

The field of AI-driven office productivity has transitioned from basic text processing to the evaluation of autonomous agents capable of handling complex, multi-step data workflows.

\subsection{Spreadsheet and Tabular Manipulation}
Early efforts in evaluating spreadsheet capabilities focused on mapping natural language to specific formulas or isolated operations. InstructExcel ~\cite{singh2023instructexcelbenchmarknatural} provided an initial foundation for translating user intent into spreadsheet actions by providing a dataset of natural language instructions paired with their corresponding Excel executions. More recently, SpreadsheetBench ~\cite{tang2024spreadsheetbenchtowardschallenging}  shifted the focus toward real-world complexity by sourcing tasks from professional forums like ExcelForum and MrExcel. This benchmark requires models to perform sophisticated data cleaning and structural transformations, moving beyond simple arithmetic to test a model's understanding of spreadsheet logic and cell dependencies. Furthermore, the Alpha Excel Benchmark ~\cite{noever2025alphaexcelbenchmark} introduced high-difficulty challenges derived from the Financial Modeling World Cup (FMWC). These tasks are designed to evaluate the limits of LLMs in high-stakes financial environments where multi-step reasoning and precise formula composition are paramount. While these benchmarks test spreadsheet proficiency, AIDABench extends this scope by integrating these capabilities into a broader workflow that includes cross-format data intelligence and professional-grade charting.

\subsection{Document Understanding and PDF Analytics}
Data analytics in office settings often requires extracting and reasoning over information locked in unstructured or semi-structured formats. DocVQA ~\cite{Mathew2020DocVQAAD} established the standard for visual question answering on document images, containing 50,000 questions that require models to bridge the gap between computer vision and natural language processing. Building on this, OmniDocBench ~\cite{zhang2024omnidocbenchbenchmarkingdiverse} highlighted the challenges of parsing complex layouts, such as nested tables and multi-column reports, which are critical precursors to effective retrieval-augmented generation (RAG) and downstream office analysis. AIDABench builds upon these foundations by requiring models to not only extract data from heterogeneous sources like financial reports and operational records but also to synthesize that information for rigorous numerical and statistical reasoning, reflecting the actual demands of enterprise data workflows.

\subsection{Agentic Data Intelligence and Professional Workflows}
Recent research has increasingly focused on the "agentic" nature of data work, where AI must plan and execute long-horizon tasks. DAComp ~\cite{leiWedDAComp} evaluates agents across the entire data pipeline, from engineering to insight generation. DSBench ~\cite{yu2025dsbenchhowfar} evaluates agents on their ability to handle the entire data science pipeline, including data preparation, modeling, and evaluation. It emphasizes the need for agents to reason over realistic, large-scale datasets rather than toy examples. Similarly, InsightBench ~\cite{sahu2025insightbenchevaluatingbusinessanalytics} focuses on the generation of business insights from structured data. Workplace-centric benchmarks like APEX-Agents ~\cite{vidgenTue} and OdysseyBench ~\cite{xu2025odysseybenchevaluatingllm} simulate the high-pressure environments of investment banking and management consulting.

Despite these efforts, there remains a gap between isolated task performance and real-world analytical scenarios that require multi-step reasoning and domain knowledge. \textbf{Our work differs from prior benchmarks in three key aspects:} (1) we focus on end-to-end data analytics tasks rather than isolated subtasks; (2) we include diverse real-world datasets across multiple domains; and (3) we develop specialized evaluators for three task categories, designed to closely reflect human judgment.

\section{Methods }

\label{sec:headings}

\subsection{Dataset Construction}
\label{sec:dataset construction}

\begin{figure} 
    \centering
    \includegraphics[width=\textwidth]{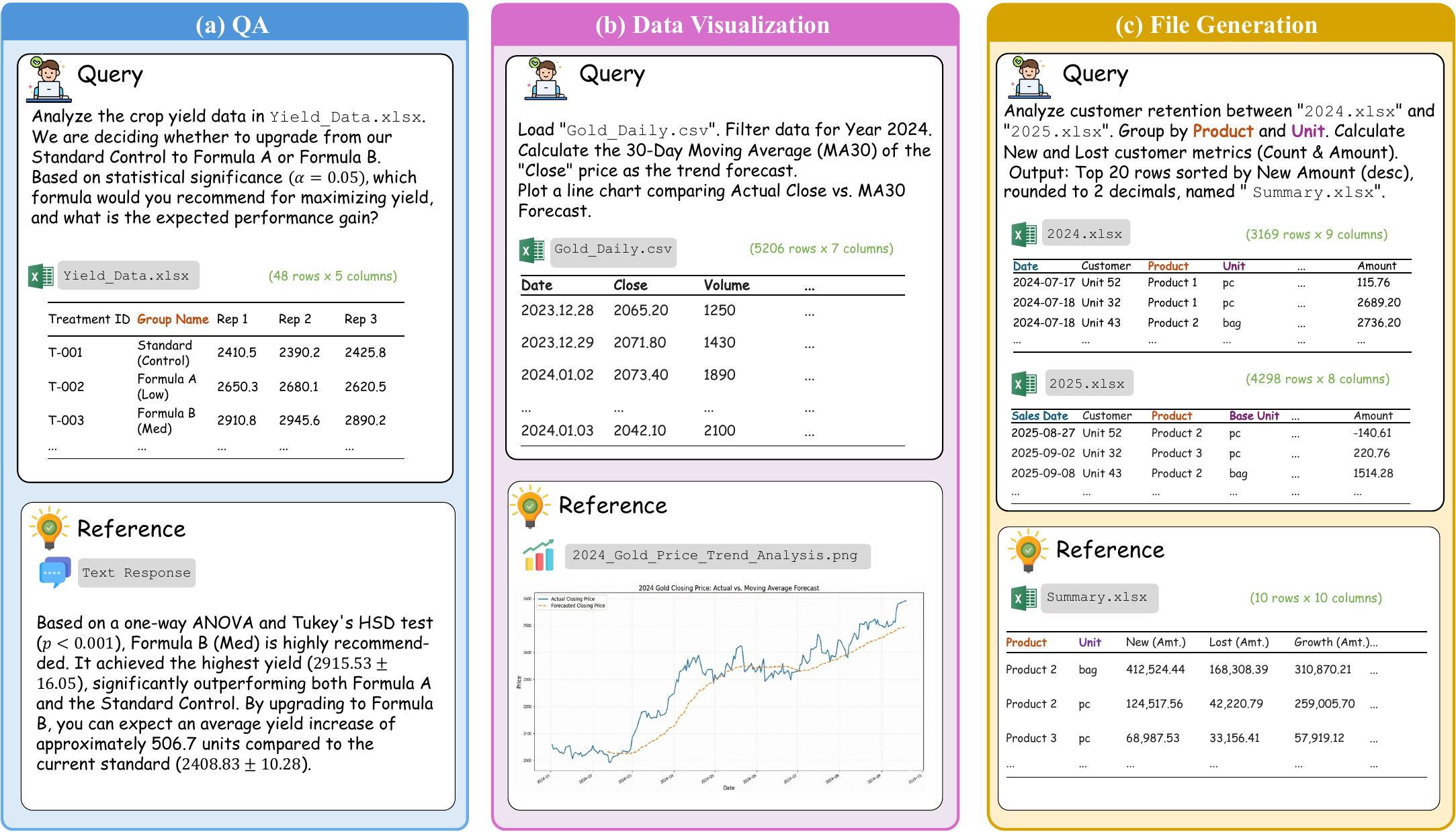}
    \caption{Three example evaluation scenarios in the benchmark. (a) QA: answer users’ data analysis questions based on the provided data; (b) Data Visualization: create visualizations based on users’ questions and the provided data. Each scenario shows the corresponding Query and the expected Reference output format; (c) File Generation: generate spreadsheets according to users’ requirements.}
    \label{fig:example}
\end{figure}
\wenxiuedit{In this section, we describe the construction of AIDABench, including data source collection, task design, annotation pipeline, and quality control procedures.}
\subsubsection{Task Design and Quality Control}
\label{sec:Task Design and Quality Control}
\wenxiuedit{In this section, we describe the data construction process of AIDABench. We assembled a team of domain users with diverse, real-world data processing and analysis needs. These users were tasked with contributing authentic analytical scenarios drawn from their daily workflows, covering domains such as finance, sales, human resources, and project management. All collected data underwent rigorous privacy protection, including the anonymization of personally identifiable information and the desensitization of business-sensitive fields, thereby grounding the benchmark in authentic use cases while ensuring compliance with data privacy regulations. 
The tasks encompass a broad range of analytical operations, including data structure recognition, data cleaning, conditional filtering, group aggregation, sorting, ranking, and descriptive statistics. To ensure benchmark quality, the ground-truth answer for each instance is independently labeled and cross-reviewed by at least two additional experts to minimize subjective bias, with annotation and validation typically requiring $1$–$2$ hours per task.}

\subsubsection{Dataset Statistics}
\label{sec:dataset_statistics}

\textbf{Output modalities and data type coverage.} The dataset comprises three primary task categories: question answering (QA) ($37.5\%$), data visualization ($19.2\%$), and file generation ($43.3\%$). A representative example for each category is shown in Fig.~\ref{fig:example}. Regarding input document formats, tabular files dominate the distribution (xlsx/csv: $91.8\%$), complemented by additional formats (e.g., docx and pdf) to support mixed-type processing.

\textbf{Task complexity stratification.} We measure task complexity by the number of expert-level reasoning steps required to reach the final answer. Tasks are categorized into three difficulty levels: easy ($\leq 6$ steps, $29.5\%$), medium ($7$–$12$ steps, $49.4\%$), and hard ($\geq 13$ steps, $21.1\%$). Notably, medium and hard tasks collectively account for over $70\%$ of the dataset, ensuring sufficient representation of challenging, multi-step analytical scenarios.

\textbf{Scale and cross-file complexity.} The benchmark comprises $603$ tasks in total. Among them, $27.4\%$ are cross-file tasks requiring joint reasoning over multiple input files, with a single task involving up to $14$ files. These tasks evaluate the model's ability to align, integrate, and reconcile information across heterogeneous sources. Specifically, $16.6\%$ of tasks involve exactly two files, while $10.8\%$ involve three or more.

\subsection{Inference Protocol}
\label{sec:inference}

All models are evaluated under a unified, tool-augmented protocol designed to reflect real-world document analysis scenarios. For each task instance, the model receives the task instruction along with the associated input file(s), and is prompted to decompose the request into sub-tasks while iteratively validating intermediate results. The complete prompt template is provided in Appendix~\ref{app:task-execution-prompt}.

\textbf{Tool interface and sandbox.} Each model is provided with a single tool-call interface, \texttt{execute\_code}, capable of executing arbitrary Python code for data inspection, transformation, and output generation. All invocations are dispatched to a stateless, containerized sandbox with a pre-configured environment including commonly used data-processing libraries (e.g., \texttt{pandas}, \texttt{matplotlib}, \texttt{openpyxl}).  No state is preserved across invocations, ensuring strict execution isolation and full reproducibility.

\textbf{Execution loop and termination.} The model iteratively writes and executes Python code to (i) carry out required operations such as data inspection, cleaning, aggregation, visualization, and file generation, and (ii) verify intermediate results against the task requirements. We formalize this as a \emph{``plan--execute--verify''} loop: the model continues issuing \texttt{execute\_code} calls until all task objectives are satisfied, subject to a maximum of $20$ interaction rounds. If the model produces a final response or deliverable before reaching this limit, the run terminates early. In either case, the output is forwarded to the corresponding evaluator for scoring.

Please note that this protocol serves as a standardized reference setting for fair and reproducible comparisons, rather than a required or exclusive inference scheme. Models may be evaluated under alternative prompting strategies, agent designs, tool interfaces, or verification loops, as long as the configuration is explicitly specified; unless otherwise stated, results in this paper use the protocol described above.

\subsection{Evaluator Design}
\label{sec:evaluator-design}
To align with the three task categories in AIDABench—question answering,  data visualization, and file generation—we design a dedicated evaluator for each. All evaluators share a common input schema: the original task instruction paired with a set of verifiable key points derived from the ground-truth annotation. Each evaluator produces a structured, machine-readable judgment, enabling automated score aggregation across all task instances. An overview of the evaluation framework is illustrated in Fig.~\ref{fig:evaluator_design}.


\newlength{\evalha}
\newlength{\evalhb}
\newlength{\evalgap}
\newlength{\evalH}

\setlength{\evalha}{0.14\textheight} 
\setlength{\evalhb}{0.18\textheight} 
\setlength{\evalgap}{0.8em}          
\setlength{\evalH}{\dimexpr \evalha + \evalhb + \evalgap\relax} 

\begin{figure*}[t]
  \centering

  \begin{minipage}[t]{0.48\textwidth}\vspace{0pt}
    \centering

    \begin{subfigure}[t]{\linewidth}
      \centering
      \makebox[\linewidth][c]{%
        \includegraphics[height=\evalha]{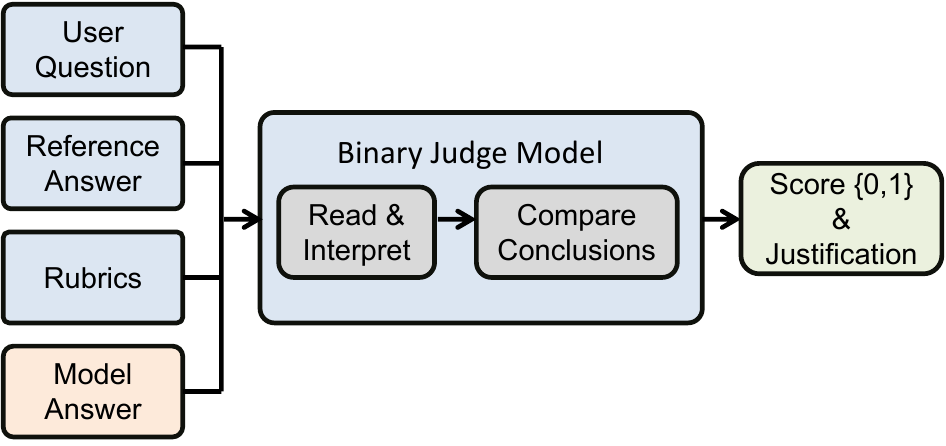}%
      }
      \caption{QA Evaluator}
      \label{fig:qa_evaluator}
    \end{subfigure}

    \vspace{\evalgap}

    \begin{subfigure}[t]{\linewidth}
      \centering
      \makebox[\linewidth][c]{%
        \includegraphics[height=\evalhb]{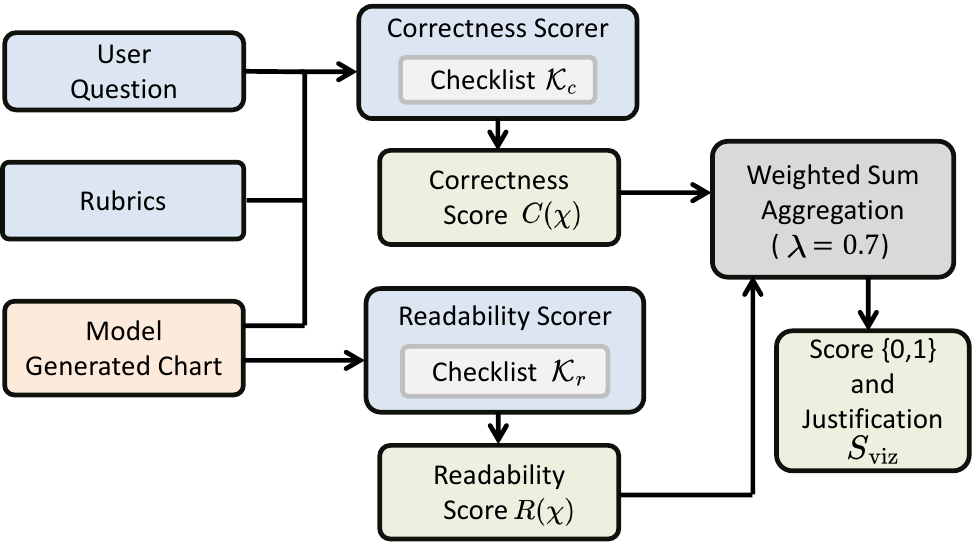}%
      }
      \caption{Visualization Evaluator}
      \label{fig:chart_evaluator}
    \end{subfigure}
  \end{minipage}
  \hfill
  \begin{minipage}[t]{0.48\textwidth}\vspace{0pt}
    \centering
    \begin{subfigure}[t]{\linewidth}
      \centering
      \makebox[\linewidth][c]{%
        \includegraphics[height=\evalH]{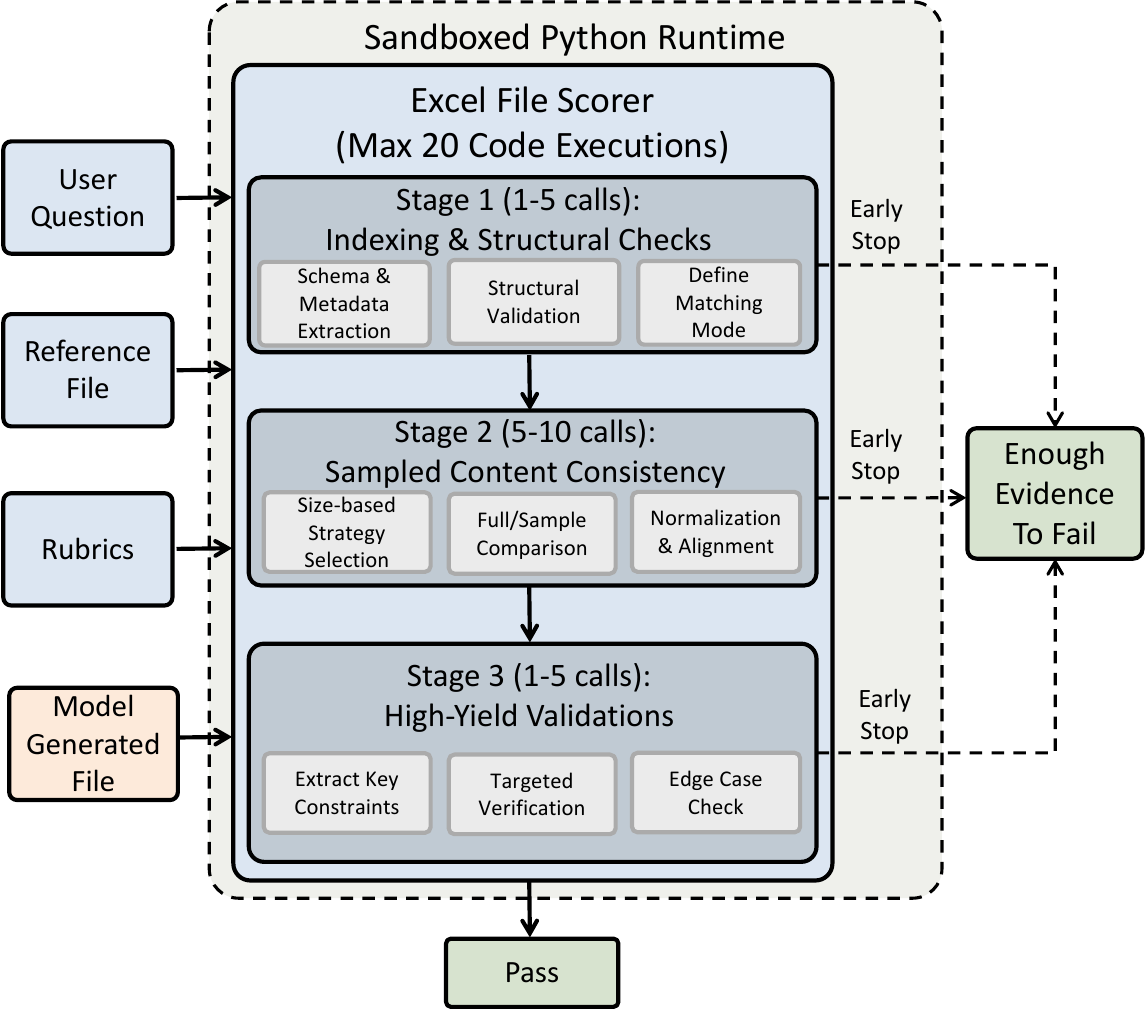}%
      }
      \caption{Spreadsheet File Evaluator}
      \label{fig:excel_evaluator}
    \end{subfigure}
  \end{minipage}

  \caption{The design of three types of evaluators in AIDABench.}
  \label{fig:evaluator_design}
\end{figure*}


\subsubsection{QA Evaluator}
\label{sec:qa-evaluator}
We score QA tasks using a binary judge (Fig.~\ref{fig:qa_evaluator}) prompted with the \emph{question}, \emph{reference answer}, and (if provided)
\emph{key points}. The judge follows a short, deterministic workflow: (i) read and interpret the question and
reference answer; (ii) compare the model's final conclusion against the reference under the question context; and
(iii) assign a score in $\{0,1\}$ with a brief justification. For quantitative/statistical questions, the judge
requires strict agreement on reported values, allowing equivalent rounding when applicable. For knowledge and
analysis questions, the judge checks whether the response covers the reference's core claims and satisfies the key
points. The complete prompt template is provided in Appendix~\ref{app:answer-scoring-prompt}.

\subsubsection{Visualization Evaluator}
\label{sec:viz-evaluator}
We evaluate visualization tasks with two complementary judges (Fig.~\ref{fig:chart_evaluator}): a \emph{correctness} scorer and a \emph{readability}
scorer. Both follow a rubric checklist protocol: for each item, the judge assigns $1$ if it is directly
verifiable from the rendered chart and satisfied, and 0 otherwise (including non-verifiable cases). The overall
workflow is: (i) score correctness key points; (ii) score readability key points; (iii) convert both to normalized
rates and aggregate them into a single score. Formally, for a chart $\chi$ with correctness checklist
$\mathcal{K}_{c}$ and readability checklist $\mathcal{K}_{r}$, we compute
\begin{equation}
C(\chi)=\frac{1}{|\mathcal{K}_{c}|}\sum_{k\in\mathcal{K}_{c}}\mathbf{1}\!\left[\operatorname{Sat}(\chi,k)\right],
\qquad
R(\chi)=\frac{1}{|\mathcal{K}_{r}|}\sum_{k\in\mathcal{K}_{r}}\mathbf{1}\!\left[\operatorname{Sat}(\chi,k)\right],
\end{equation}
where $\operatorname{Sat}(\chi,k)$ is true if rubric item $k$ is satisfied and observable from the chart. The final
visualization score is a weighted sum:
\begin{equation}
S_{\text{viz}}(\chi)=\lambda\,C(\chi)+(1-\lambda)\,R(\chi),\qquad \lambda=0.7.
\end{equation}
The prompt templates for the correctness and readability judges are included in
Appendix~\ref{app:viz-scoring-prompts}.

\subsubsection{Spreadsheet File Evaluator}
\label{sec:excel-file-evaluator}
For file-generation tasks, we design a Spreadsheet File Evaluator (Fig.~\ref{fig:excel_evaluator}) that assesses whether a model-produced spreadsheet satisfies the task requirements by comparing it against the reference file in light of the original task instruction. 
At each stage, the evaluator terminates immediately upon finding sufficient evidence of failure;  otherwise, it advances to the next stage. If no critical violation is detected by the end, the output is marked \textit{Pass}.

\textbf{Stage~1: Structural Indexing and Validation.}
The evaluator first constructs a structural index for both the reference and predicted files, covering sheet names, dimensions, column headers, and data types. When the task prescribes an exact output structure, strict matching is enforced and any mismatch triggers immediate rejection. Otherwise, the evaluator falls back to a tolerant mode that accommodates benign discrepancies such as row/column permutations, extra columns, equivalent nulls, and minor numeric deviations ($\varepsilon{=}10^{-6}$). A set of reusable sample positions is also pre-selected to avoid redundant scans in subsequent stages.

\textbf{Stage~2: \wenxiuedit{Sampled Content Verification.}}
Based on sheet size, the evaluator chooses a full comparison for small sheets and a sampled comparison for large ones. Before comparison, it normalizes values (numeric
tolerance, canonical dates, trimmed strings, unified nulls) and aligns columns by name under the chosen matching mode.
It then checks whether the sampled contents are consistent; significant discrepancies trigger early rejection.

\textbf{Stage 3: \wenxiuedit{Task-Specific Constraint Validation.}}
Finally, the evaluator extracts question-relevant constraints from the instruction/key points and runs targeted checks
with high payoff, such as deduplication policy (e.g., keep-first), filters, sorting, group-by aggregates, Top-$K$,
derived columns, and key statistics. It may also probe a few edge cases. If no critical violations are found within the
budget, the file is marked \textit{Pass}; otherwise it returns a concise failure rationale.

\wenxiuedit{The tool-call budget for evaluators is distributed across the three stages as follows: up to 5 invocations for structural validation, up to 10 for content verification, and up to 5 for constraint validation, totaling at most 20 invocations per task.} The complete evaluator prompt is provided in Appendix~\ref{app:file-judge-prompt}.

\subsubsection{Evaluator Selection and Calibration}
\label{sec:evaluator-calibration}

Each evaluator is powered by a carefully selected LLM: qwq-32B~\cite{qwen_qwq32b_2025} for the QA Evaluator, gemini-3-pro-preview~\cite{google_gemini3_2025} for the Visualization Evaluator, and claude-sonnet-4-5~\cite{anthropic_claude_sonnet_45_2025} for the Spreadsheet File Evaluator. To validate reliability, we conduct a calibration study comparing evaluator outputs against expert annotations. The QA and Spreadsheet File Evaluators both achieve over $95\%$ agreement with human judges, while the Visualization Evaluator exceeds $90\%$.

\section{Results }
\label{sec:results}

\subsection{Experimental Setup}
\label{subsec:eval_setting_brief}

\paragraph{Decoding Hyperparameters.}

To ensure a fair and reproducible comparison, we refrain from task-specific tuning of decoding hyperparameters. Instead, we adopt each provider's recommended temperature setting for the corresponding model, along with the associated sampling strategy (e.g., Top-$p$, Top-$k$) when explicitly documented. For endpoints where certain parameters are unsupported (e.g., some reasoning-mode variants), we defer to provider defaults to avoid unintended behavior. Full per-model decoding configurations are reported in Appendix~\ref{app:decoding_settings} (Table~\ref{tab:recommended_decoding_settings}).

\paragraph{Protocol.}
All models are evaluated using the same tool-augmented pipeline with a maximum of 20 \texttt{execute\_code} invocations per instance. Unless otherwise stated, an auxiliary spreadsheet summary (detailed in  Appendix ~\ref{app:aux-spreadsheet-summarization}) is provided as supplementary context in the full setting. We report both pass@1 (single run) and pass@3 (best of three runs) for the main results. For data visualization tasks, where the evaluator yields a soft score in $[0,1]$, the pass@3 score is computed by taking the maximum score across three runs per instance and then averaging over all instances. Ablation experiments are repeated three times and reported as the mean (avg@3) to provide a stable estimate of each component's contribution.

\paragraph{Additional metrics.}
We additionally track token consumption and the number of interaction rounds per run, with avg@3 statistics reported in Appendix~\ref{subsec:efficiency}.

\paragraph{Performance by difficulty.}
We further break down results by task difficulty (as defined in Section~\ref{sec:dataset_statistics});
the full score distributions are provided in Appendix~\ref{app:difficulty}.

\subsection{Overall Model Performance}
\label{subsec:overall}

\begin{table*}[t]
\centering
\small
\setlength{\tabcolsep}{6pt}
\renewcommand{\arraystretch}{1.15}
\resizebox{\textwidth}{!}{%
\begin{tabular}{l l c c c c c c c c}
\toprule
& & \multicolumn{4}{c}{\textbf{pass@3}} & \multicolumn{4}{c}{\textbf{pass@1}} \\
\cmidrule(lr){3-6}\cmidrule(lr){7-10}
\textbf{Model} & \textbf{Model Size} &
\textbf{Overall} & \textbf{QA} & \textbf{Data Vis.} & \textbf{File Gen.} &
\textbf{Overall} & \textbf{QA} & \textbf{Data Vis.} & \textbf{File Gen.} \\
\midrule
claude-sonnet-4-5 & Undisclosed
& $\mathbf{70.78}$ & $\mathbf{72.12}$ & $78.28$ & $\mathbf{66.28}$
& $\mathbf{59.43}$ & $\mathbf{68.58}$ & $\mathbf{67.71}$ & $\mathbf{49.43}$ \\

gemini-3-pro-preview & Undisclosed
& $69.25$ & $71.24$ & $\mathbf{78.97}$ & $63.22$
& $57.39$ & $64.60$ & $63.87$ & $48.28$ \\

qwen3-max-2026-01-23-thinking & Undisclosed
& $66.34$ & $71.68$ & $75.05$ & $57.85$
& $56.67$ & $66.81$ & $64.41$ & $44.83$ \\

deepseek-v3.2-thinking & $685\mathrm{B}$
& $65.72$ & $69.47$ & $71.83$ & $59.77$
& $53.81$ & $62.39$ & $60.83$ & $43.30$ \\

deepseek-v3.2 & $685\mathrm{B}$
& $65.47$ & $69.91$ & $75.68$ & $57.09$
& $51.35$ & $60.18$ & $62.51$ & $40.61$ \\

qwen3-max-2026-01-23 & Undisclosed
& $63.37$ & $69.03$ & $73.41$ & $54.02$
& $51.52$ & $60.62$ & $63.53$ & $39.08$ \\

kimi-k2-thinking & $1\mathrm{T}$ ( $32\mathrm{B}$ active )
& $62.24$ & $69.47$ & $64.90$ & $54.79$
& $54.16$ & $66.37$ & $59.97$ & $41.00$ \\

qwen3-235b-a22b-instruct-2507 & $235\mathrm{B}$ ( $22\mathrm{B}$ active )
& $61.23$ & $65.04$ & $76.91$ & $50.96$
& $47.80$ & $53.98$ & $63.15$ & $36.02$ \\

kimi-k2-turbo-preview & $1\mathrm{T}$ ( $32\mathrm{B}$ active )
& $55.56$ & $62.83$ & $67.25$ & $44.06$
& $38.25$ & $46.90$ & $48.93$ & $27.59$ \\

qwen3-30b-a3b-thinking-2507 & $30\mathrm{B}$ ( $3\mathrm{B}$ active )
& $48.18$ & $60.62$ & $62.53$ & $31.03$
& $38.66$ & $52.65$ & $50.94$ & $22.22$ \\

qwen3-30b-a3b-instruct-2507 & $30\mathrm{B}$ ( $3\mathrm{B}$ active )
& $43.45$ & $48.67$ & $61.22$ & $31.03$
& $31.37$ & $35.40$ & $47.57$ & $21.46$ \\
\bottomrule
\end{tabular}%
}
\caption{Performance on AIDABench under pass@3 and pass@1.
\textbf{Data Vis.} denotes \emph{Data Visualization}, and \textbf{File Gen.} denotes \emph{File Generation}.
Model sizes are reported when available; otherwise marked as \emph{Undisclosed}.
\textbf{Best} results in each column are highlighted in bold.}
\label{tab:aidabench_results}
\end{table*}

Table~\ref{tab:aidabench_results} reports overall and per-category results on AIDABench under both pass@3 and pass@1. Even the best-performing model achieves only $70.78$ (pass@3) and $59.43$ (pass@1), underscoring the substantial challenge posed by end-to-end document data analysis.

Among the evaluated models, claude-sonnet-4-5 achieves the strongest overall performance, ranking first on both pass@3 ($70.78$) and pass@1 ($59.43$). It also leads in QA ($72.12$ / $68.58$) and File Generation ($66.28$ / $49.43$). gemini-3-pro-preview follows as a close runner-up overall ($69.25$ / $57.39$) and attains the highest Data Visualization score ($78.97$ pass@3), demonstrating particular strength on visualization-centric tasks.

Task difficulty varies notably across categories. Data Visualization yields the highest scores for most models, often exceeding $70$ (pass@3) for top systems, whereas File Generation is consistently the most challenging, with the best pass@1 reaching only $49.43$. The gap between pass@3 and pass@1 is particularly pronounced for File Generation, suggesting that multiple attempts can improve success rates on file-producing tasks but still leave a substantial margin for improvement.

More broadly, model capacity correlates with performance: smaller models (e.g., qwen3-30b-a3b-instruct-2507) consistently trail their larger counterparts, with the most pronounced deficits on File Generation. Overall, these results confirm that AIDABench meaningfully differentiates current models across diverse, realistic data-analysis workloads.

Beyond task accuracy, we analyze token consumption and interaction rounds as efficiency proxies. As detailed in Appendix~\ref{subsec:efficiency}, File Generation generally demands more rounds and tokens than QA and Data Visualization, reflecting the higher computational cost of structured file editing.


\subsection{Ablation Study}
\label{subsec:ablation-summary}

\begin{table*}[t!]
\centering
\small
\setlength{\tabcolsep}{6pt}
\renewcommand{\arraystretch}{1.15}
\resizebox{\textwidth}{!}{%
\begin{tabular}{l c c c c c c c c}
\toprule
\textbf{Model} &
\textbf{Overall} & \textbf{$\Delta$} &
\textbf{QA} & \textbf{$\Delta$} &
\textbf{Data Vis.} & \textbf{$\Delta$} &
\textbf{File Gen.} & \textbf{$\Delta$} \\
\midrule
claude-sonnet-4-5
& $58.62$ & \down{0.84} & $68.14$ & \down{3.39} & $65.63$ & \up{1.67} & $47.25$ & \up{0.26} \\

gemini-3-pro-preview
& $56.37$ & \up{0.38} & $61.95$ & \up{2.01} & $64.45$ & \down{0.36} & $47.95$ & \down{0.70} \\

qwen3-max-2026-01-23-thinking
& $55.95$ & \down{1.41} & $65.63$ & \down{1.32} & $62.68$ & \up{2.73} & $44.57$ & \down{3.32} \\

deepseek-v3.2-thinking
& $52.12$ & \down{0.51} & $60.32$ & \down{0.14} & $58.62$ & \down{0.12} & $42.15$ & \down{1.03} \\

deepseek-v3.2
& $50.99$ & \up{0.25} & $58.7$ & \up{0.59} & $62.17$ & \up{2.75} & $39.34$ & \down{1.15} \\

qwen3-max-2026-01-23
& $50.26$ & \down{0.85} & $59.23$ & \down{2.45} & $58.51$ & \up{0.62} & $38.83$ & \down{0.13} \\

kimi-k2-thinking
& $48.72$ & \up{2.12} & $62.6$ & \down{1.54} & $58.3$ & \up{3.11} & $37.16$ & \up{0.13} \\

qwen3-235b-a22b-instruct-2507
& $46.86$ & \down{6.30} & $53.69$ & \down{6.20} & $60.24$ & \down{4.28} & $34.99$ & \down{7.28} \\

kimi-k2-turbo-preview
& $37.32$ & \down{3.14} & $45.28$ & \up{0.30} & $46.34$ & \down{7.17} & $26.44$ & \down{4.35} \\

qwen3-30b-a3b-thinking-2507
& $38$ & \down{10.37} & $51.5$ & \down{10.05} & $48.41$ & \down{16.84} & $21.29$ & \down{7.37} \\

qwen3-30b-a3b-instruct-2507
& $28.89$ & \down{6.00} & $31.27$ & \down{9.00} & $43.87$ & \down{7.64} & $20.18$ & \down{2.68} \\
\bottomrule
\end{tabular}%
}

\caption{Ablation on auxiliary spreadsheet summary (\textbf{w/} vs.\ \textbf{w/o}).
For each setting, all scores are reported as \textbf{avg@3} (mean over three independent runs).
The table shows the \textbf{w/} scores, and $\Delta$ denotes the change when removing the summary, i.e.,
$\Delta = \text{score(w/o summary)} - \text{score(w/ summary)}$. $\uparrow$ denotes improvement and $\downarrow$ denotes degradation.
\textbf{Data Vis.} = Data Visualization; \textbf{File Gen.} = File Generation.}

\label{tab:ablation_wo_info_delta_overall}
\end{table*}
\wenxiuedit{We investigate the effect of providing the auxiliary spreadsheet summary by comparing two settings: with and without the summary, while holding all other variables (inference protocol, tool budget, and evaluators) fixed.  Each configuration is evaluated over three independent runs, and we report avg@3 (mean over three runs). The performance difference is computed as 
$\Delta=\text{score (w/o summary)}-\text{score (w/ summary)}$ (Table~\ref{tab:ablation_wo_info_delta_overall}). Note that the main leaderboard uses pass@3 (best of three) to reflect retry-assisted performance, whereas ablations adopt avg@3 to provide a more stable estimate of each component's effect. We choose avg@3 here because the max operator amplifies variance and can obscure consistent performance shifts, making $\Delta$ unreliable as an effect-size estimate.
We observe that removing the summary degrades performance for the majority of models (8 out of 11), with the largest drops observed on smaller models. Breaking down by category, the removal most consistently hurts QA, followed by File Generation, whereas Data Visualization shows no clear trend across models.}



\subsection{Error Analysis}
\label{subsec:Bad case study}

\begin{figure} 
    \centering
    \includegraphics[width=\textwidth]{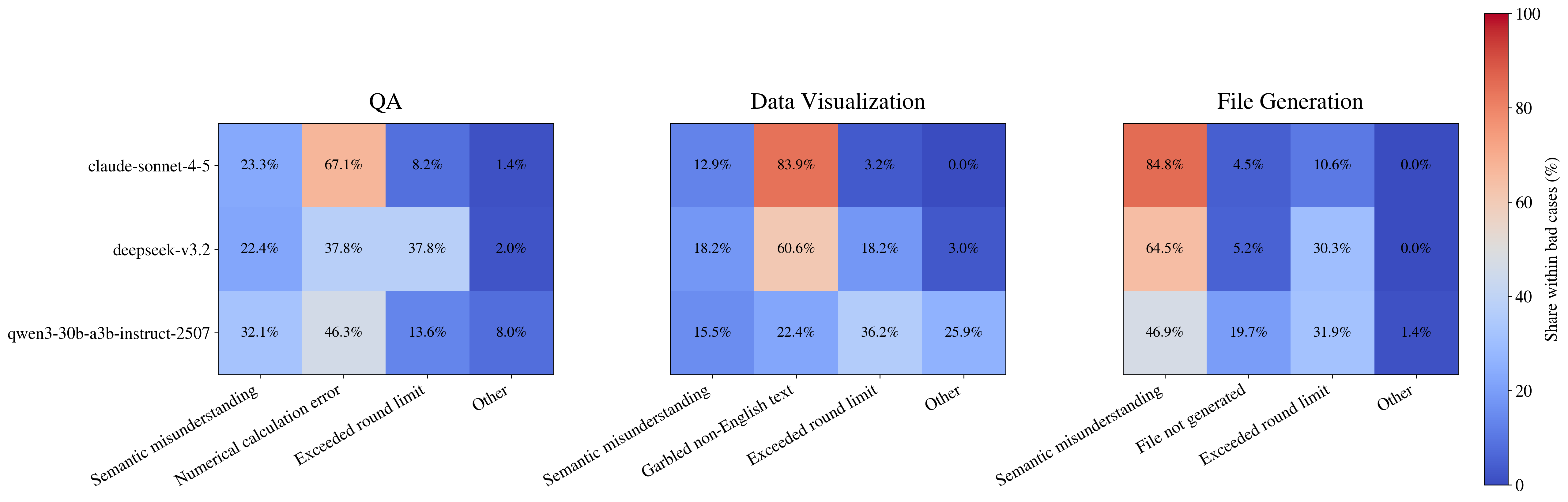}
    \caption{Error composition within bad cases across scenarios. Heatmaps report, for each model and scenario (QA, Data Visualization, and File Generation), the percentage share of each error type among the bad cases. Redder colors indicate higher shares. Model names are shown on the left panel, and a single color scale is shared across all panels.}

        \label{fig:badcase_matrix}
\end{figure}

We selected claude-sonnet-4-5, deepseek-v3.2, and qwen3-30b-a3b-instruct-2507 as representative models with strong, medium, and weak performance, respectively, to assess error composition across different capability levels, as detailed in Fig.~\ref{fig:badcase_matrix}.

\noindent\textbf{QA.} Numerical calculation errors constitute the primary failure mode for claude-sonnet-4-5 ($67.1\%$). deepseek-v3.2 faces significant convergence challenges, as evidenced by the high prevalence of exceeded round limits ($37.8\%$). qwen3-30b-a3b-instruct-2507 exhibits compound weaknesses, significantly impacted by both calculation errors ($46.3\%$) and semantic misunderstandings ($32.1\%$).

\noindent\textbf{Data Visualization.} 
The errors of claude-sonnet-4-5 are overwhelmingly due to garbled non-English text rendering ($83.9\%$). deepseek-v3.2 shares this deficiency ($60.6\%$) but also frequently hits the round limit ($18.2\%$). qwen3-30b-a3b-instruct-2507 exhibits a more dispersed error profile, dominated by round-limit exceedances ($36.2\%$) and miscellaneous failures ($25.9\%$).

\noindent\textbf{File Generation.} Semantic misunderstanding constitutes the leading error category for all three models ($84.8\%$, $64.5\%$, and $46.9\%$, respectively), yet deepseek-v3.2 and qwen3-30b-a3b-instruct-2507 are further bottlenecked by round-limit exceedances ($30.3\%$ and $31.9\%$), suggesting convergence difficulties that do not arise in claude-sonnet-4-5.

Representative qualitative examples are provided in Appendix~\ref{app:badcase}.

\section{Future Works}
\label{sec:future_works}

Building on the current benchmark and evaluation pipeline, several extensions can further increase coverage and
practicality. First, we will broaden task diversity toward longer-horizon, multi-file workflows and richer
document layouts, while maintaining the same end-to-end framing across QA, data visualization, and file generation.
Second, we will strengthen robustness testing, including more systematic stress cases for formatting fidelity (e.g.,
multilingual text rendering in plots) and distribution shifts that mirror real deployments. Finally, we will improve
the scalability of evaluation: although the current judge-and-verify setup provides reliable supervision, the scoring
models can be compute- and token-intensive. Future iterations will explore cost-efficient hybrids (deterministic checks
plus lightweight learned scorers), judge distillation, and reuse of verification artifacts to reduce overhead without
sacrificing calibration quality.

\section{Conclusion}
\label{sec:conclusion}
We presented AIDABench, a benchmark for end-to-end data analytics over realistic office documents, spanning QA,
data visualization, and file generation under a unified tool-augmented inference protocol. We also developed
task-aligned evaluators that enable automated and fine-grained assessment across the three categories. Extensive
experiments on a diverse set of state-of-the-art models demonstrate clear headroom on complex, multi-step workflows,
with file generation remaining especially challenging. Beyond accuracy, our results and analysis highlight efficiency
as a first-class consideration for practical analytics agents, motivating future work on more scalable, cost-effective
evaluation and inference. Overall, AIDABench offers a rigorous foundation for measuring progress, diagnosing failure
modes, and guiding the development of more reliable data analytics systems.

\bibliographystyle{unsrt}  
\bibliography{references}  

\clearpage
\appendix
\section*{Appendix}
\addcontentsline{toc}{section}{Appendix}

\section{Task Execution Prompt}
\label{app:task-execution-prompt}

We use the following prompt template to instruct the model to act as a data analysis assistant that solves the given task by analyzing the provided file(s), decomposing complex requests into manageable sub-tasks, and iteratively validating intermediate results under a strict tool-call budget.

\begin{lstlisting}[style=promptstyle,language=Python,caption={Prompt template for task execution with tool-call budget.}]
You are a data analysis assistant. You will complete the given task by analyzing a file.

The current task is:
{task_prompt}
Never forget this task!

### Note: The overall task may be very complex!
Please split the task into multiple sub-tasks, then execute them step by step, and finally complete the task.

Below are some important tips that may help you solve the task:

#### <tips>
- You need to continuously run code via tool calls until the task is completed. However, you must return the final result within 20 rounds; otherwise you will exceed the round limit and trigger an exception!
- You do not need to install new Python packages; the environment already has the required packages installed.
- If one approach fails, try another. A correct answer exists!
- When the question mentions "the N-th column", use code to verify the actual position of the N-th column.
- Each tool call runs in a fresh sandbox environment. This means you must write complete code each time, including all necessary imports.
- Always verify that your final answer is correct!
- If you encounter errors, try debugging your code.
- Carefully consider whether the task requires deduplication (removing duplicates).
- If you see paths like /mnt/data or /mnt/result, they are virtual paths and will be replaced at runtime. Therefore, do not try to inspect files using a bare "mnt/" path, because "mnt/" alone will not be recognized/replaced and may cause errors.
- If a tool call fails, or the code does not run correctly, do NOT assume the returned result is correct and continue reasoning based on it.
  The correct approach is to analyze the error cause and try to fix it.
- If your obtained result does not satisfy the task, keep analyzing and iterating until your final answer satisfies all requirements.
\end{lstlisting}

\section{QA Scoring Prompt}
\label{app:answer-scoring-prompt}

We use the following prompt template to score an AI model's answer against the question and reference answer, producing a binary score (0/1) with a brief justification.

\begin{lstlisting}[style=promptstyle,language=Python,caption={Prompt template for answer scoring.}]
You are a scoring assistant. Please score the AI model's answer. We will provide:
1. The question and the reference answer
2. The AI model answer to be scored

Scoring rubric: There are only two possible scores: 0 or 1. Based on the question and the reference answer, judge the AI model's answer as follows:
1. Quantitative/statistics questions: If the AI model's computed result is basically consistent with the reference answer (statistical values must match more strictly), assign 1; if there is a clear discrepancy between the computed result and the reference answer, assign 0.
2. Knowledge Q&A and analysis questions: If the AI model's answer fully covers the rubric items of the reference answer and its logic matches the original meaning, assign 1; if the answer significantly deviates from the reference answer or misses rubric items, assign 0.

Notes:
1. The above is the general scoring rubric. However, some questions provide [rubrics]. If [rubrics] are provided, you must consider them when scoring.
2. The model answer may include reasoning steps; focus primarily on the final conclusion.
3. Numeric values may keep different numbers of decimals; if they match after rounding, consider them correct.
4. When comparing tables, carefully understand the question first and then compare.
5. Focus only on the answer logic; any language is acceptable (Chinese, English, ...).

Finally, follow these steps to score:
1. Understand the question and the reference answer.
2. Score the AI model's answer according to the rubric above, and explain why.
3. Output using the following JSON format:

{"score": <0/1>, "reason": "[Explain why you gave this score]"}

We will provide the user's question, the reference answer, and the AI model answer.
<Question>:
{question}
<Reference Answer>:
{reference}
<Rubrics>:
{rubrics}
<AI Model Answer>:
{AI_answer}
\end{lstlisting}

\section{Visualization Scoring Prompts}
\label{app:viz-scoring-prompts}

This appendix provides the prompts used to score data-visualization outputs. We use two complementary judges:
(1) a \textit{chart correctness scorer} that verifies task-specific rubrics, and
(2) a \textit{chart readability scorer} that evaluates visual presentation quality.

\subsection{Chart Correctness Scoring Prompt}
\label{app:chart-correctness-prompt}

\begin{lstlisting}[style=promptstyle,language=Python,caption={Prompt template for chart correctness scoring.}]
You are a "Chart Correctness Scorer". You will be given:
- question: the user's original request
- rubrics: a list of answer rubrics (each is a verifiable checkpoint; satisfied = 1 point)
- chart: the chart to be scored (may be an image, a screenshot, or a textual description/specification)

Your task: determine whether the chart satisfies each item in rubrics, and compute the total score.

Scoring rules:
1) Each number of rubric items must be scored as either 0 or 1: satisfied = 1; not satisfied OR not verifiable from the chart = 0.
2) You must judge ONLY based on information visible in the chart. Do not guess or hallucinate. If it is not observable, count it as not satisfied.
3) This score evaluates data correctness, NOT visual aesthetics.
4) total_number_of_rubric_items = number of items in rubrics; score = number of satisfied rubrics.
5) The reason must be concise and should summarize the main causes of point deductions.

Inputs:
question: {question}
rubrics: {rubrics}
chart: see the image below.

Output requirements (VERY IMPORTANT):
- Output JSON only. Do not output any extra text. Do not use markdown.
- The JSON object must contain ONLY the following fields:
  {{
    "correct_number_of_rubric_items": <int>,
    "total_number_of_rubric_items": <int>,
    "reason": "<string>"
  }}
\end{lstlisting}

\subsection{Chart Readability Scoring Prompt}
\label{app:chart-readability-prompt}

\begin{lstlisting}[style=promptstyle,language=Python,caption={Prompt template for chart readability scoring.}]
You are a "Chart Readability Scorer". You will be given a single chart image.
Your task: evaluate the chart's readability and presentation quality according to the dimensions below.
This evaluation focuses on whether the chart is easy to read and interpret, NOT whether it is visually beautiful.

Dimensions:
1. labels_and_titles: Clear, concise, and correctly placed.
2. layout_spacing: Well-organized and not cluttered.
3. color_accessibility: Distinct and colorblind-friendly where possible.
4. axis_scaling: Axes labeled correctly with proportional scaling.
5. chart_type_suitability: Appropriate chart type for the data/task.
6. font_and_legends: Readable fonts and properly aligned legends.
7. annotation_readability: Data labels/annotations/callouts/leader lines are clear, appropriately placed, and non-overlapping.
8. visual_hierarchy_and_emphasis: The key takeaway is immediately apparent; comparisons are highlighted effectively, and secondary information is appropriately de-emphasized (avoiding distraction from the main message).

Scoring rules:
1) Each dimension is scored as 0 or 1: 1 = good/satisfied; 0 = poor/not satisfied/not verifiable.
2) Provide a brief reason for each dimension.

Output requirements (VERY IMPORTANT):
- Output JSON only. Do not output any extra text. Do not use markdown.
- The JSON object must contain the following fields:
  {{
    "labels_and_titles": <0/1>,
    "labels_and_titles_reason": "<string>",
    "layout_spacing": <0/1>,
    "layout_spacing_reason": "<string>",
    "color_accessibility": <0/1>,
    "color_accessibility_reason": "<string>",
    "axis_scaling": <0/1>,
    "axis_scaling_reason": "<string>",
    "chart_type_suitability": <0/1>,
    "chart_type_suitability_reason": "<string>",
    "font_and_legends": <0/1>,
    "font_and_legends_reason": "<string>",
    "annotation_readability": <0/1>,
    "annotation_readability_reason": "<string>",
    "visual_hierarchy_and_emphasis": <0/1>,
    "visual_hierarchy_and_emphasis_reason": "<string>"
  }}
\end{lstlisting}

\section{Spreadsheet Judge Prompt}
\label{app:file-judge-prompt}

We provide the full prompt used for our \textit{File Judge} (evaluation agent) to determine whether a model-generated file satisfies the task requirements by comparing it against a reference file. The agent is constrained by a strict tool-call budget and follows a coarse-to-fine, sampling-first verification strategy.

\begin{lstlisting}[style=promptstyle,language=Python,caption={Prompt template for the Spreadsheet Judge evaluation agent.}]
You are an intelligent file evaluator (File Judge).
Goal: Given the "question + reference file", decide whether the "prediction file" satisfies the requirements, and output a JSON verdict.

Question: {{question}}

Files:
- Reference file: "{{reference_path}}"
- Prediction file: "{{prediction_path}}"

[Tool-Call Budget (MUST FOLLOW)]
- You may call `execute_code` at most 20 times (never exceed 20).
- Before each tool call, explicitly state in text what you are going to verify, to avoid aimless reruns.
- Once you have sufficient evidence, you MUST stop immediately and output the final JSON.

[Overall Strategy: Coarse-to-Fine, Sampling-First, Deepen Only If Necessary]
Follow the 3 stages below and stop early whenever possible:

Stage 1 (1-5 calls): Quick probing / indexing
- Read basic metadata of both files: file type, sheet list, row/column counts per sheet, column-name sets, and dtype overviews.
- If the question specifies strict requirements on header/column order/output format/file structure, enforce strict matching:
  * If any mismatch is found, output conclusive evidence for `false` and stop.
  Otherwise, use a relaxed comparison:
  * allow row/column reordering,
  * allow extra columns (as long as they do not affect the requested outputs),
  * allow small floating-point differences (tol = 1e-6),
  * treat different missing-value representations as equivalent (NaN / None / "" are considered equal),
  * verify content consistency via sampling (head / middle / tail / random).
- In code, pre-generate reusable sample row indices (head/middle/tail + random) to avoid rescanning.

Stage 2 (5-10 calls): Sample-based content consistency checks
(Do NOT do full row-by-row comparison if the file is large; if small, full comparison is allowed.)
- For each comparable sheet, perform a sampled comparison:
  * first N rows + last N rows + middle N rows + random N rows
  * default N = 30; if the sheet has fewer rows, shrink N automatically or compare all rows.
- Compare after normalization (ignore small float errors, ignore row/column order):
  * Columns: align by column names
  * Rows: use "row hashing" or "sort + align" for set-like comparison (samples only)
  * Numbers: abs/rel tol = 1e-6
  * Dates: convert to ISO format
  * Strings: strip whitespace
  * Missing values: unify representations
- If sampled differences are significant, output `false` with evidence and stop.
  If samples match, proceed to Stage 3.

Stage 3 (use only 1-5 additional calls): Task-relevant key validations only
- Extract only high-yield constraints from the question (avoid heavy NLP):
  Examples: deduplication/keep-first, sorting, filtering conditions, group-by aggregation,
  Top-K, new headers/new columns, presence of specific values, and key statistics
  (sum/mean/count), etc.
- Validate constraints that can be computed quickly.

[Prohibited Behaviors (VERY IMPORTANT)]
- Do NOT perform full row-by-row comparison on large files unless:
  * the file is small (e.g., < 200 rows), or
  * the question explicitly requires exact full-file equality.
- Do NOT keep calling tools "to verify more details". Never exceed 20 calls.
- Do NOT output anything other than the final JSON.
  (Intermediate results should only be computed internally in code; do not print.)

[Final Output (output exactly once)]
Output a single JSON object:
{{"is_correct": true/false,
 "reason": "Concise explanation: which sheets were checked, sampling strategy, and key evidence of mismatch or key checks passed."}}

Now begin. You may call `execute_code`, but you MUST strictly follow the staged strategy and budget.

\end{lstlisting}

\section{Decoding Hyperparameters}
\label{app:decoding_settings}

We report the decoding hyperparameters used in our experiments in
Table~\ref{tab:recommended_decoding_settings}.
Our primary goal is to minimize confounding factors introduced by ad-hoc decoding choices.
Therefore, we adopt the \emph{provider-recommended} temperature for each model and only set
additional sampling parameters (Top-$p$, Top-$k$) when they are explicitly specified as part
of the recommended preset. When a parameter is not supported (or is documented to be ignored),
we keep it unspecified and use provider defaults.

\begin{table*}[t]
\centering
\small
\setlength{\tabcolsep}{6pt}
\renewcommand{\arraystretch}{1.15}
\resizebox{\textwidth}{!}{%
\begin{tabular}{l c c c l}
\toprule
& \multicolumn{3}{c}{\textbf{Recommended Sampling}} & \\
\cmidrule(lr){2-4}
\textbf{Model} &
\textbf{Temp.} & \textbf{Top-$p$} & \textbf{Top-$k$} &
\textbf{Notes} \\
\midrule
claude-sonnet-4-5
& $0.0$ & -- & -- & Use either Temp. or Top-$p$ (do not set both). \\
gemini-3-pro-preview
& $1.0$ & $0.95$ & $40$ & Use provider suggested sampling preset. \\
qwen3-max-2026-01-23-thinking
& $0.6$ & $0.95$ & $20$ & Thinking preset. \\
deepseek-v3.2-thinking
& -- & -- & -- & Thinking mode ignores sampling params (use defaults). \\
deepseek-v3.2
& $1.0$ & $1.0$ & -- & Provider default sampling. \\
qwen3-max-2026-01-23
& $0.7$ & $0.8$ & $20$ & Instruct preset. \\
kimi-k2-thinking
& $1.0$ & default & default & Provider recommended temperature. \\
qwen3-235b-a22b-instruct-2507
& $0.7$ & $0.8$ & $20$ & Instruct preset. \\
kimi-k2-turbo-preview
& $0.6$ & default & default & Turbo preset. \\
qwen3-30b-a3b-thinking-2507
& $0.6$ & $0.95$ & $20$ & Thinking preset. \\
qwen3-30b-a3b-instruct-2507
& $0.7$ & $0.8$ & $20$ & Instruct preset. \\
\bottomrule
\end{tabular}%
}
\caption{Recommended decoding hyperparameters per model. We follow each provider's recommended temperature and adopt the associated sampling parameters (Top-$p$, Top-$k$) when specified. Unspecified parameters are left to provider defaults.}
\label{tab:recommended_decoding_settings}
\end{table*}

\noindent\textbf{Clarifications.}
``--'' indicates that the parameter is intentionally not set.
``default'' indicates that we do not override the parameter and use the provider default.
For models/endpoints that ignore sampling knobs in thinking mode, we keep these parameters unset.

\section{Auxiliary Spreadsheet Summarization}
\label{app:aux-spreadsheet-summarization}

\begin{figure*}[t]
  \centering
  \includegraphics[width=\textwidth]{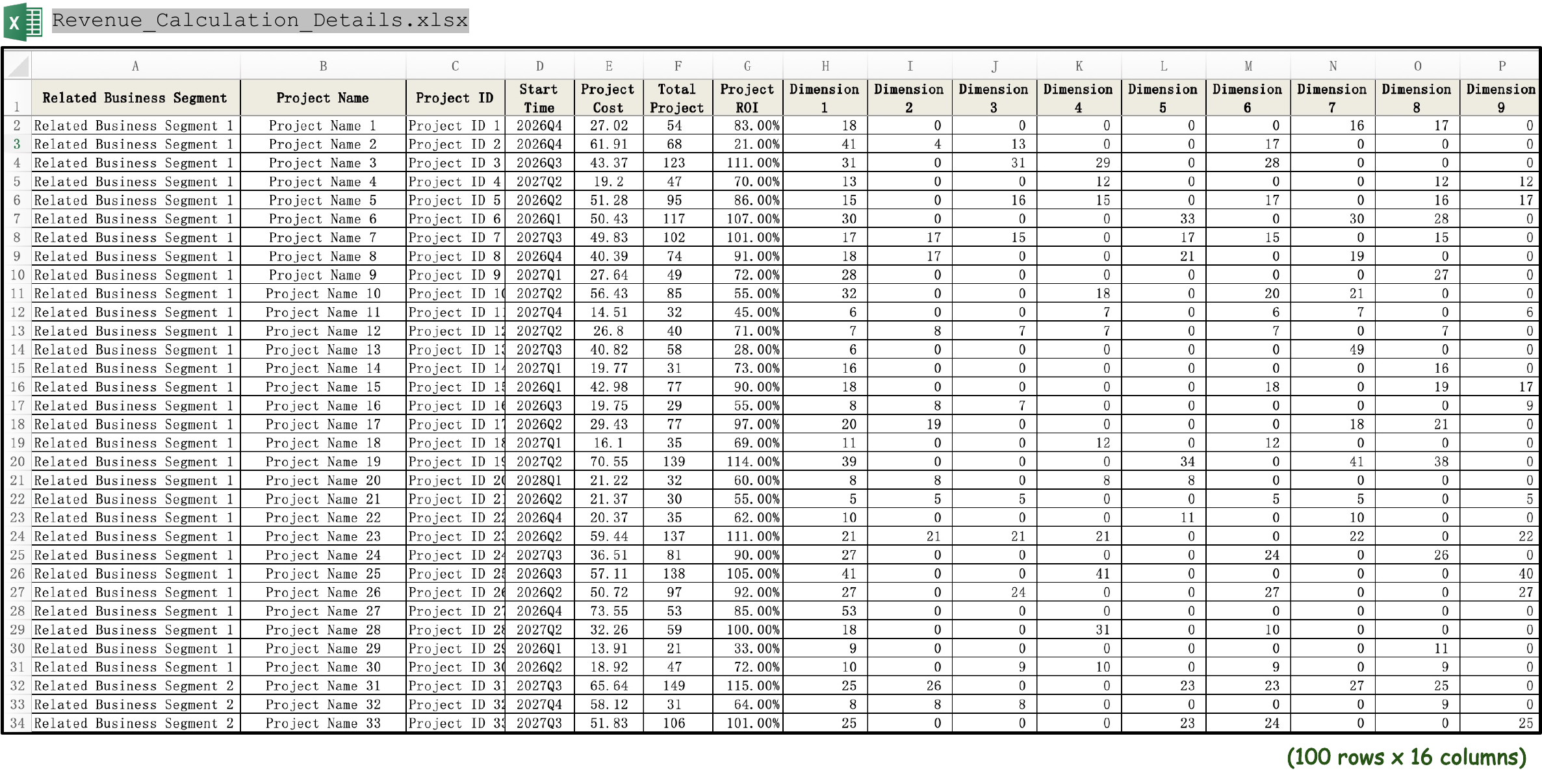}
  \caption{Example of auxiliary spreadsheet summarization input: screenshot of the workbook \texttt{Revenue\_Calculation\_Details.xlsx} (active sheet excerpt).}
  \label{fig:aux-revenue-before}
\end{figure*}

We show an example of the auxiliary spreadsheet summarization output.
Given an input workbook, the module emits a token-efficient, machine-readable summary including:
(i) workbook metadata (file type, sheet names, active sheet), and
(ii) a sparse cell map for the active sheet.
Figure~\ref{fig:aux-revenue-before} shows the input workbook excerpt, and
Listing~\ref{lst:aux-revenue-json} presents the corresponding full JSON summary emitted by the module.

\FloatBarrier

\begin{lstlisting}[style=jsonstyle,caption={Example of auxiliary spreadsheet summarization.},label={lst:aux-revenue-json}]
{
  "fileName": "Revenue_Calculation_Details.xlsx",
  "fileType": "xlsx",
  "sheetNames": ["Project Revenue Details"],
  "sheets": [
    {
      "sheetName": "Project Revenue Details",
      "sheetType": "ReadOnlyWorksheet",
      "cells": {"A1": "Related Business Segment", "B1": "Project Name", "C1": "Project ID", "D1": "Start Time", "E1": "Project Cost (10k CNY)", "F1": "Total Project Revenue (10k CNY)", "G1": "Project ROI", "H1": "Dimension 1", "I1": "Dimension 2", "J1": "Dimension 3", "K1": "Dimension 4", "L1": "Dimension 5", "M1": "Dimension 6", "N1": "Dimension 7", "O1": "Dimension 8", "P1": "Dimension 9", "A2": "Related Business Segment 1", "B2": "Project Name 1", "C2": "Project ID 1", "D2": "2026Q4", "E2": "27.02", "F2": "53.74", "G2": "0.83", "H2": "18.34", "I2": "0", "J2": "0", "K2": "0", "L2": "0", "M2": "0", "N2": "16.23", "O2": "17.17", "P2": "0", "A3": "Related Business Segment 1", "B3": "Project Name 2", "C3": "Project ID 2", "D3": "2026Q4", "E3": "61.91", "F3": "68.26", "G3": "0.21", "H3": "41.08", "I3": "4.31", "J3": "13.16", "K3": "0", "L3": "0", "M3": "16.71", "N3": "0", "O3": "0", "P3": "0", "A4": "Related Business Segment 1", "B4": "Project Name 3"}
    }
  ],
  "mode": "Delete empty cell, Omit the longer content and replace it with..., Omit lines with the same structure, Omit later lines",
  "ActiveSheet": "Project Revenue Details"
}
\end{lstlisting}

\section{Additional Experiments: Performance by Difficulty}
\label{app:difficulty}

We stratify task difficulty by operation-chain length, defined as the number of key expert operations (Steps).
Low complexity ($\leq 6$) accounts for $29.5$\% of tasks, medium ($7$--$12$) for $49.4$\%, and high ($\geq 13$) for $21.1$\%.
Table~\ref{tab:pass3_by_difficulty} reports pass@3 performance on each split.

\begin{table*}[t]
\centering
\small
\setlength{\tabcolsep}{6pt}
\renewcommand{\arraystretch}{1.15}
\begin{tabular}{l c c c}
\toprule
\textbf{Model} & \textbf{Pass@3 (Easy)} & \textbf{Pass@3 (Medium)} & \textbf{Pass@3 (Hard)} \\
\midrule
claude-sonnet-4-5 & $84.34$ & $\mathbf{71.31}$ & $\mathbf{50.54}$ \\
gemini-3-pro-preview & $\mathbf{86.97}$ & $68.30$ & $46.65$ \\
qwen3-max-2026-01-23-thinking & $82.31$ & $67.15$ & $42.08$ \\
deepseek-v3.2-thinking & $83.67$ & $66.78$ & $38.10$ \\
deepseek-v3.2 & $83.25$ & $65.92$ & $39.51$ \\
qwen3-max-2026-01-23 & $79.84$ & $64.95$ & $36.60$ \\
kimi-k2-thinking & $80.16$ & $63.27$ & $34.69$ \\
qwen3-235b-a22b-instruct-2507 & $78.52$ & $60.84$ & $37.92$ \\
kimi-k2-turbo-preview & $74.31$ & $55.51$ & $29.38$ \\
qwen3-30b-a3b-thinking-2507 & $73.79$ & $46.90$ & $15.31$ \\
qwen3-30b-a3b-instruct-2507 & $64.35$ & $44.21$ & $12.39$ \\
\bottomrule
\end{tabular}
\vspace{2pt}
\caption{Pass@3 performance stratified by task difficulty. Difficulty is defined by operation-chain length (number of key expert operations/steps): easy ($\leq 6$), medium ($7$--$12$), and hard ($\geq 13$), covering $29.5$\%, $49.4$\%, and $21.1$\% of tasks, respectively.}
\label{tab:pass3_by_difficulty}
\end{table*}

\section{Token and Round Consumption}
\label{subsec:efficiency}

\begin{figure*}[!t]
  \centering
  \includegraphics[width=\textwidth]{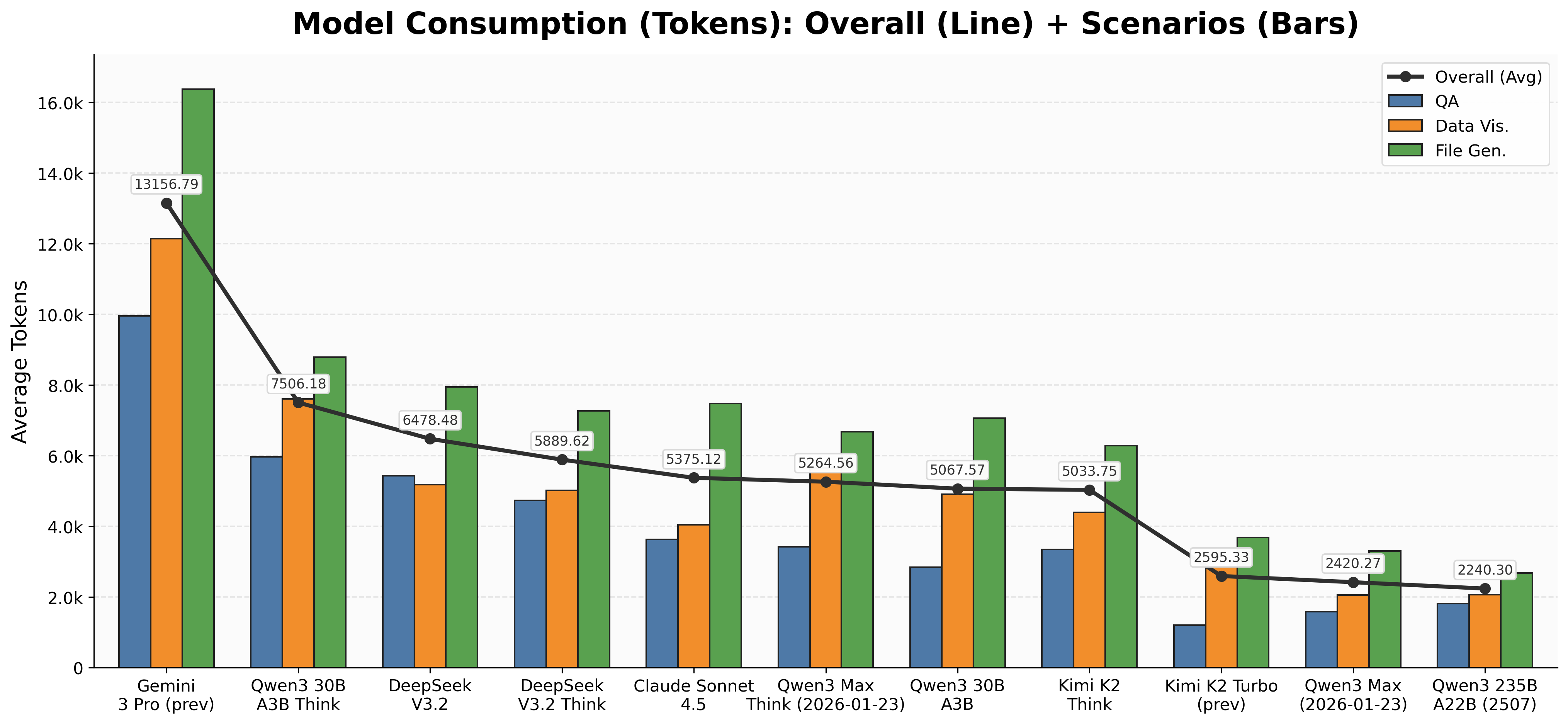}
  \caption{\textbf{Token consumption across models.} Bars show the average tokens used in each scenario (QA, Data Visualization, and File Generation), and the line shows the overall average tokens per task. Models are sorted by overall average token consumption (descending).}
  \label{fig:consumption_tokens}
\end{figure*}

\begin{figure*}[!t]
  \centering
  \includegraphics[width=\textwidth]{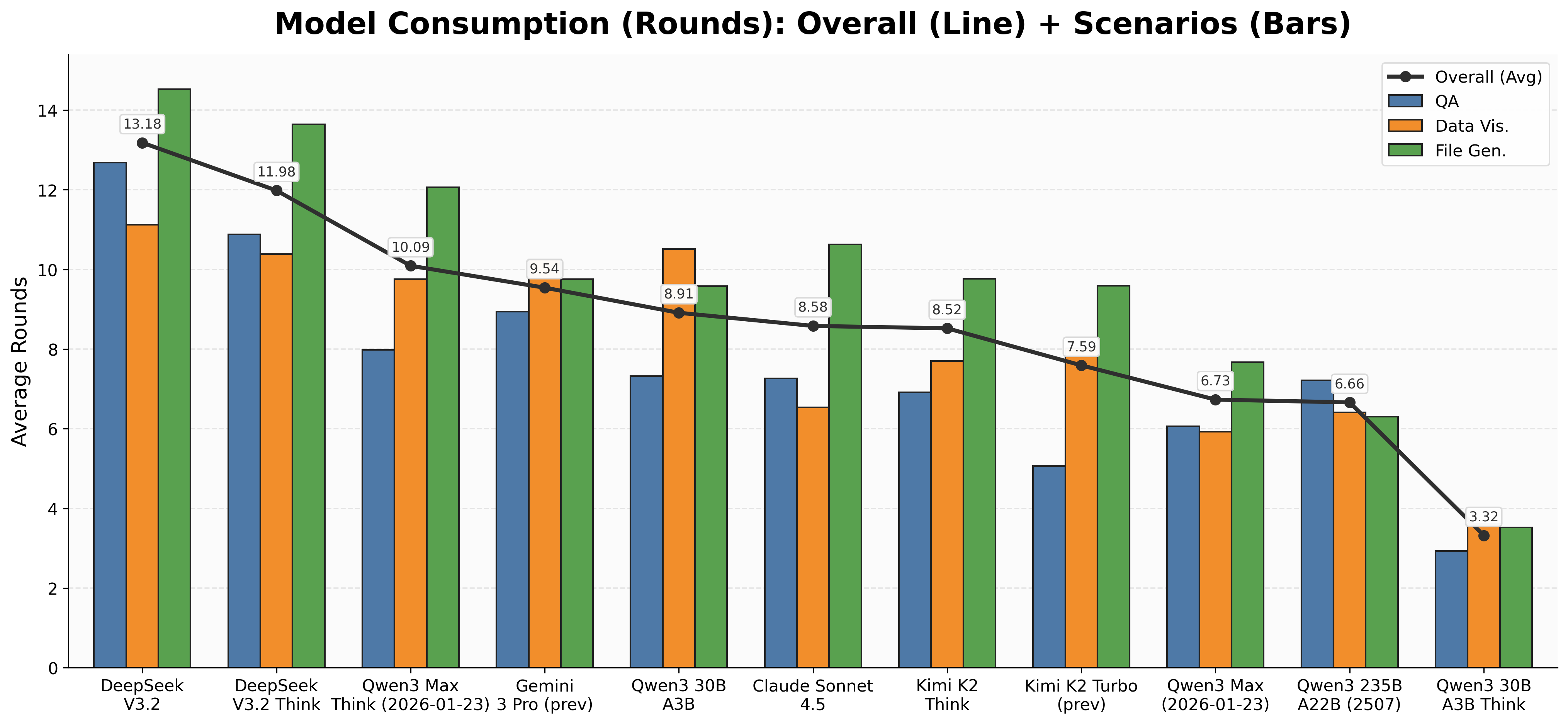}
  \caption{\textbf{Round consumption across models.} Bars show the average number of interaction rounds in each scenario (QA, Data Visualization, and File Generation), and the line shows the overall average rounds per task. Models are sorted by overall average round consumption (descending).}
  \label{fig:consumption_rounds}
\end{figure*}

Figures~\ref{fig:consumption_tokens} and~\ref{fig:consumption_rounds} summarize the average token usage and
interaction rounds per task across models. For each model, the bars report scenario-wise averages (QA, Data
Visualization, and File Generation), while the line indicates the overall average; models are sorted by overall
consumption (descending) for readability. We observe substantial efficiency variation: overall token usage ranges
from roughly $2.2$k--$13.2$k, and overall rounds range from about $3.3$--$13.2$. Across models, \emph{File Generation}
is consistently the most expensive scenario in both tokens and rounds (e.g., often requiring notably more rounds and
a larger token budget than QA and Data Visualization), reflecting the longer multi-step workflows needed to produce
structured deliverables. Token consumption also differs markedly by model behavior: gemini-3-pro-preview is
the most token-intensive overall, while qwen3-235b-a22b-instruct-2507 and qwen3-max-2026-01-23 are among the most token-efficient.
Finally, rounds and tokens do not always move together (e.g., some models use fewer rounds but longer turns),
highlighting different interaction styles and cost profiles across systems.

\section{Bad Case Examples}
\label{app:badcase}

\subsection{Semantic Misunderstanding}
\begin{figure*}[htbp]

  \centering
  \includegraphics[width=\textwidth]{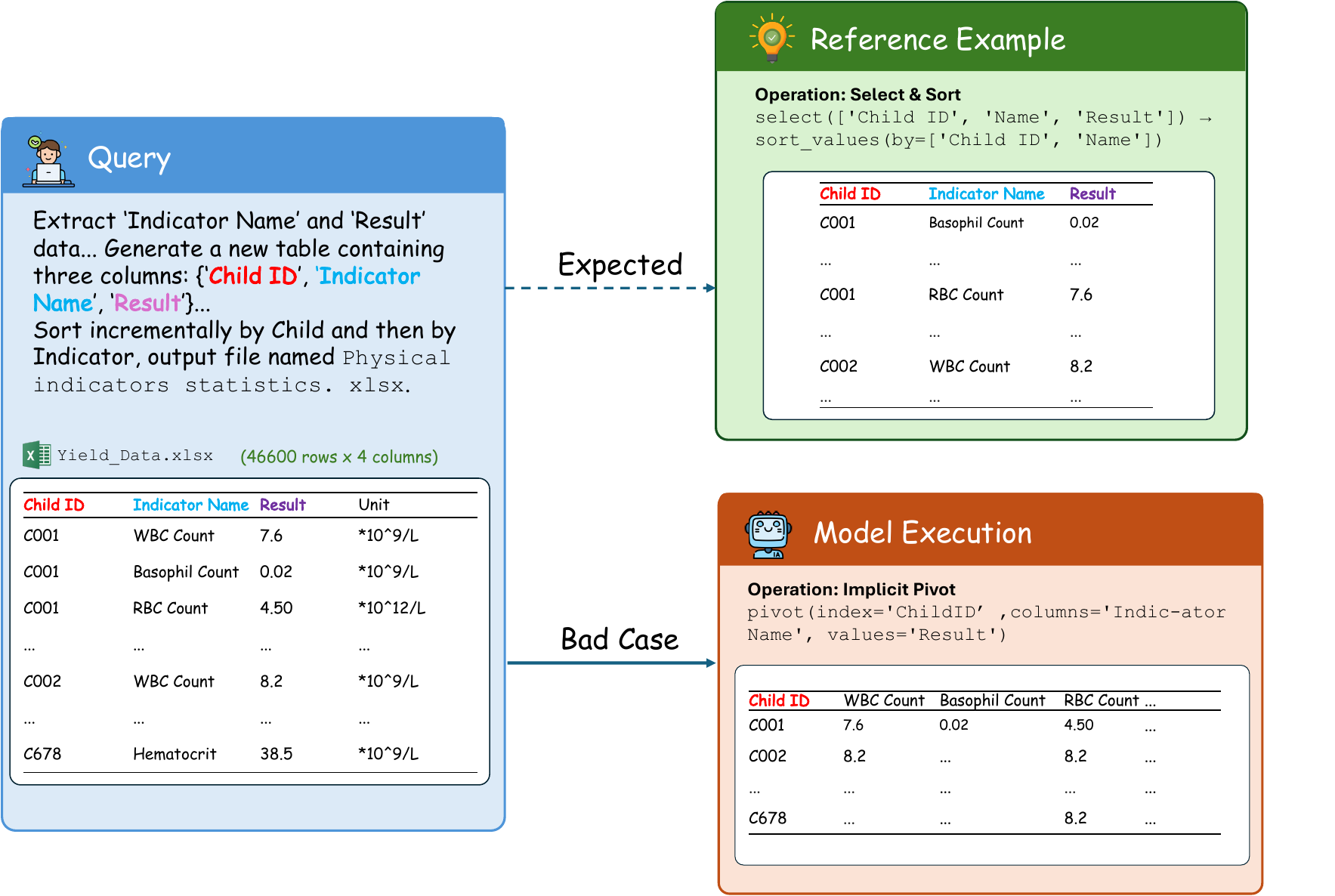}
  \caption{\textbf{Illustration of structural misalignment caused by semantic ambiguity in tabular data instructions.} The diagram contrasts the intended execution (top right), which adheres to the user's explicit "three-column" constraint using a selection and sort operation, against the model's erroneous interpretation (bottom right). The model hallucinated an unrequested pivot operation, transforming the data into a sparse wide-format matrix instead of the requested long-format tidy data.}
  \label{fig:pivot_failure}
\end{figure*}

Figure~\ref{fig:pivot_failure} shows a representative failure mode in which an LLM violates explicit output-structure constraints due to a biased interpretation of tabular semantics. 
In this example (ID: \texttt{data\_10}), the prompt clearly requires a long-format table with \textit{exactly three columns}: \{``Child ID'', ``Indicator Name'', ``Result''\}. 
The reference solution (top right) therefore applies only a projection and a row-wise sort (\texttt{select} $\rightarrow$ \texttt{sort\_values}), producing a dense $N \times 3$ DataFrame.

In contrast, the model treats the pair (``Child ID'', ``Indicator Name'') as a hierarchical key and incorrectly infers that the data should be reshaped. 
As shown in the bottom-right panel, it hallucinates an implicit pivot (i.e., an unrequested pivot-like reshape), converting indicator names into column headers and producing a sparse wide-format matrix of size roughly $N \times M$, rather than the requested $N \times 3$ output. 
This case indicates that learned priors for exploratory tabular analysis (e.g., preferring pivoted views for categorical comparison) can override explicit, user-specified requirements on table shape.

\subsection{Garbled Non-English Text in Visualizations}
\begin{figure*}[htbp]
  \centering
  \begin{subfigure}[t]{0.48\textwidth}
    \centering
    \includegraphics[width=\linewidth]{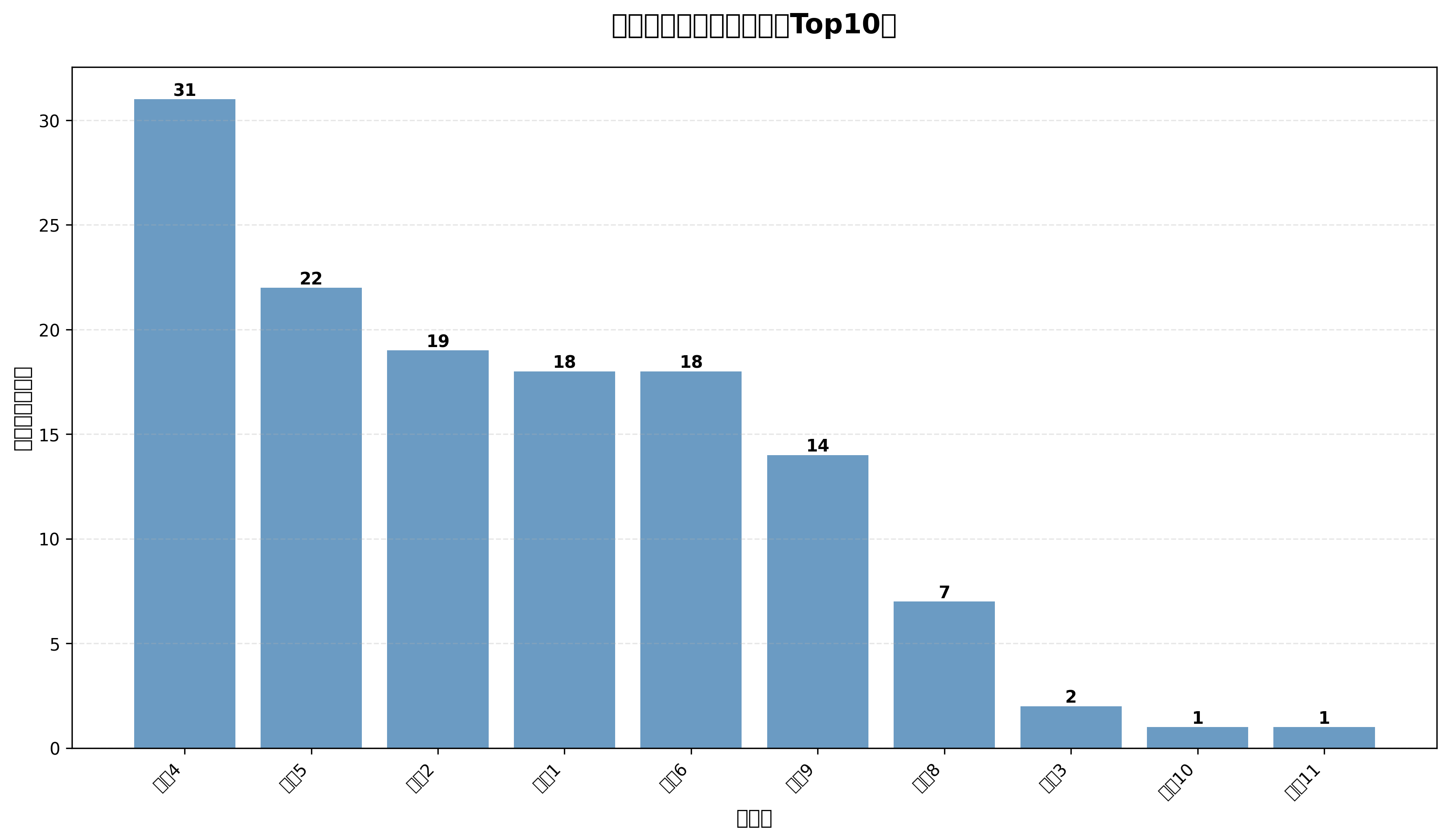}
    \caption{\textbf{Bad (garbled).} Bar chart.}
  \end{subfigure}
  \hfill
  \begin{subfigure}[t]{0.48\textwidth}
    \centering
    \includegraphics[width=\linewidth]{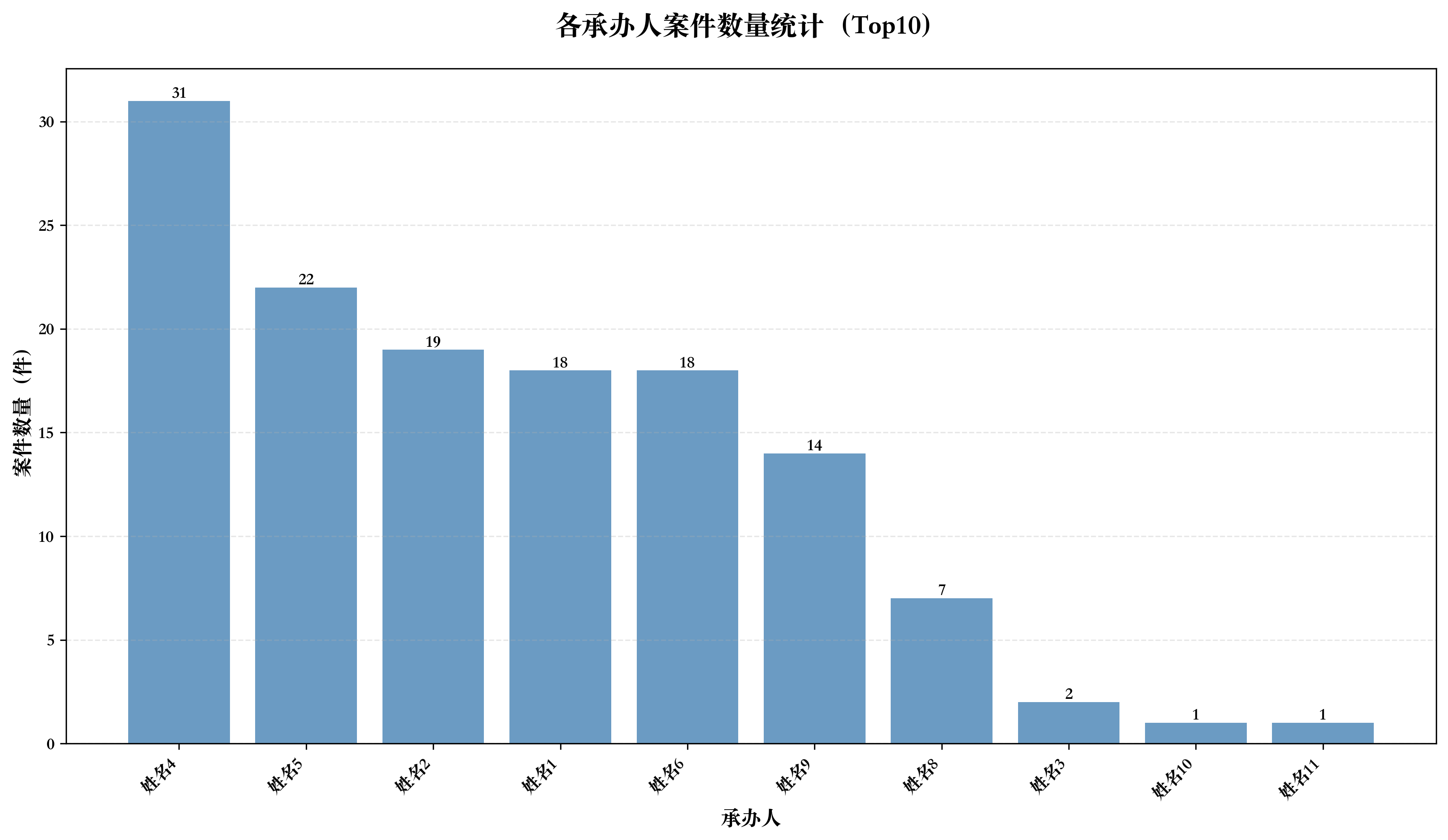}
    \caption{\textbf{Good (legible).} Bar chart.}
  \end{subfigure}

  \vspace{0.6em}

  \begin{subfigure}[t]{0.48\textwidth}
    \centering
    \includegraphics[width=\linewidth]{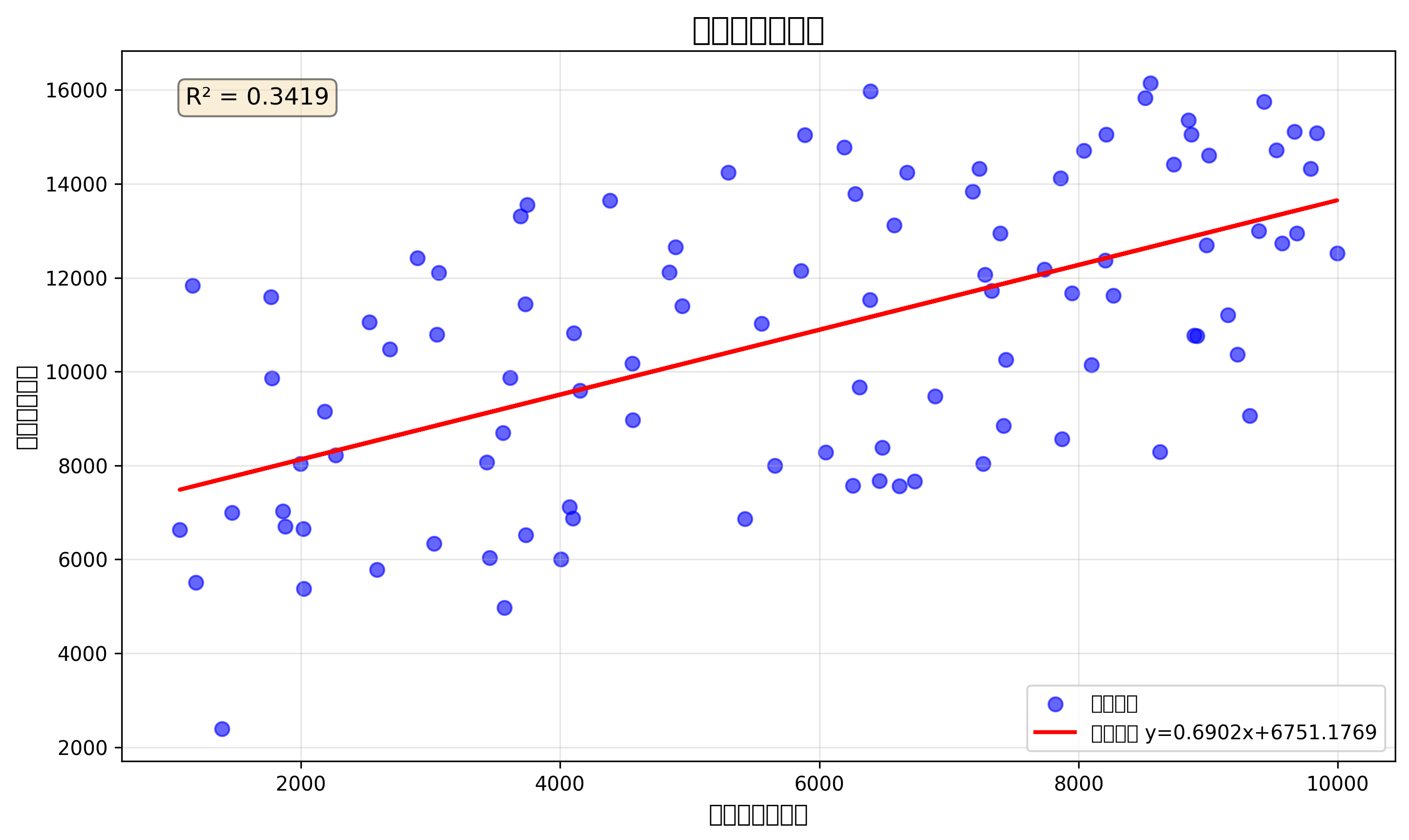}
    \caption{\textbf{Bad (garbled).}  Scatter plot with a regression line.}
  \end{subfigure}
  \hfill
  \begin{subfigure}[t]{0.48\textwidth}
    \centering
    \includegraphics[width=\linewidth]{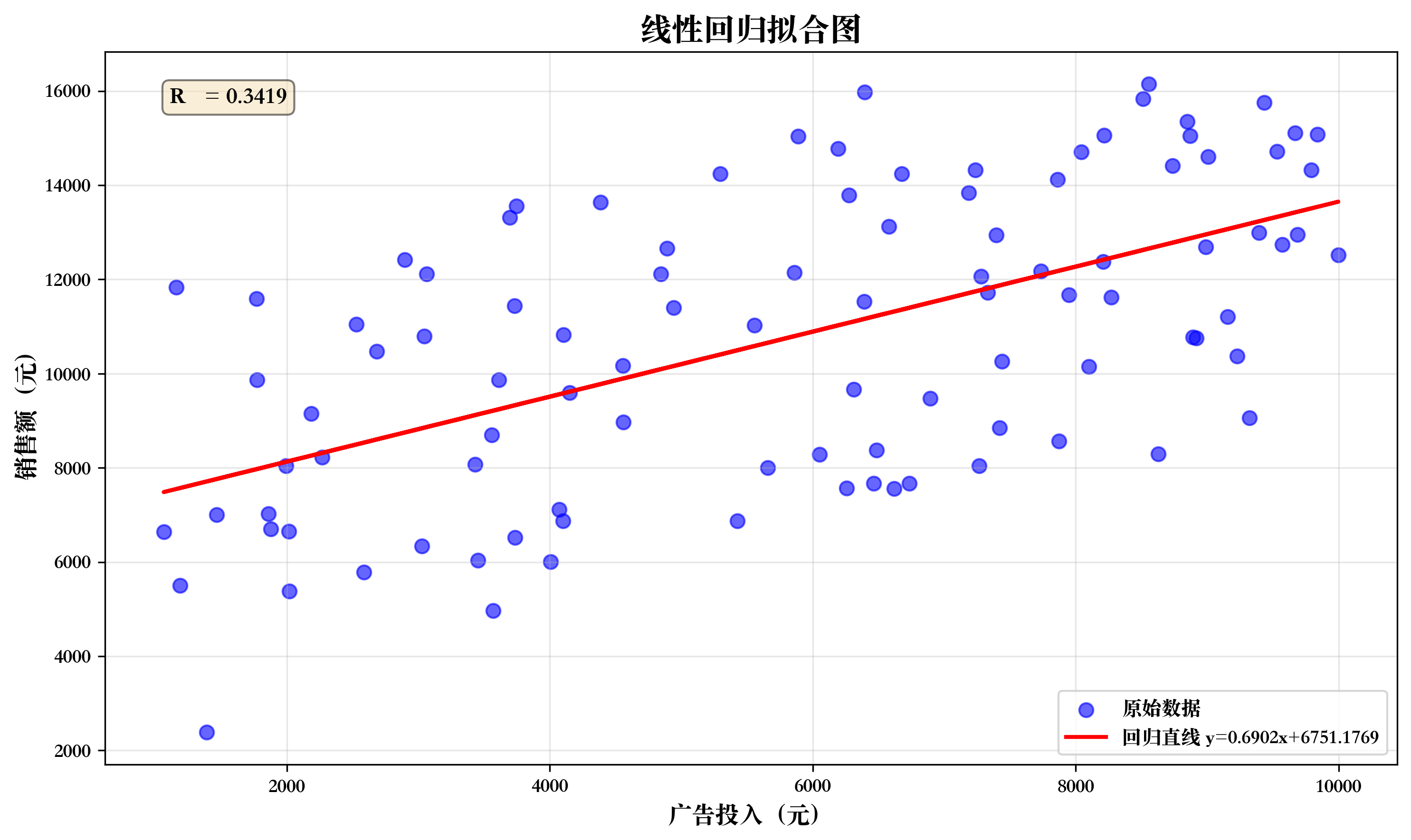}
    \caption{\textbf{Good (legible).}  Scatter plot with a regression line.}
  \end{subfigure}

  \caption{\textbf{Garbled non-English text in generated visualizations.}
  Left: the model fails to select a font with proper CJK support, causing Chinese labels to appear garbled or as
  missing-glyph placeholders. Right: with a language-compatible font, the same labels are rendered correctly,
  improving readability while keeping the chart content unchanged.}
  \label{fig:garbled_cjk}
\end{figure*}

Figure~\ref{fig:garbled_cjk} illustrates a recurring visualization failure where the model renders non-English
labels (e.g., Chinese) with incorrect or unsupported fonts. Although the plotted values and chart structure may be
correct, axis ticks, legends, or annotations become garbled (or replaced by missing-glyph symbols), substantially
reducing readability and potentially invalidating the visualization for downstream use.

This issue typically arises when the plotting backend falls back to a default font without adequate CJK glyph
coverage or fails to embed the intended font during export. In contrast, using a font family that explicitly
supports the target language produces legible text and preserves the communicative intent of the chart.

\FloatBarrier

\subsection{Numerical Calculation Error}
\begin{figure*}[htbp]
  \centering
  \includegraphics[width=\textwidth]{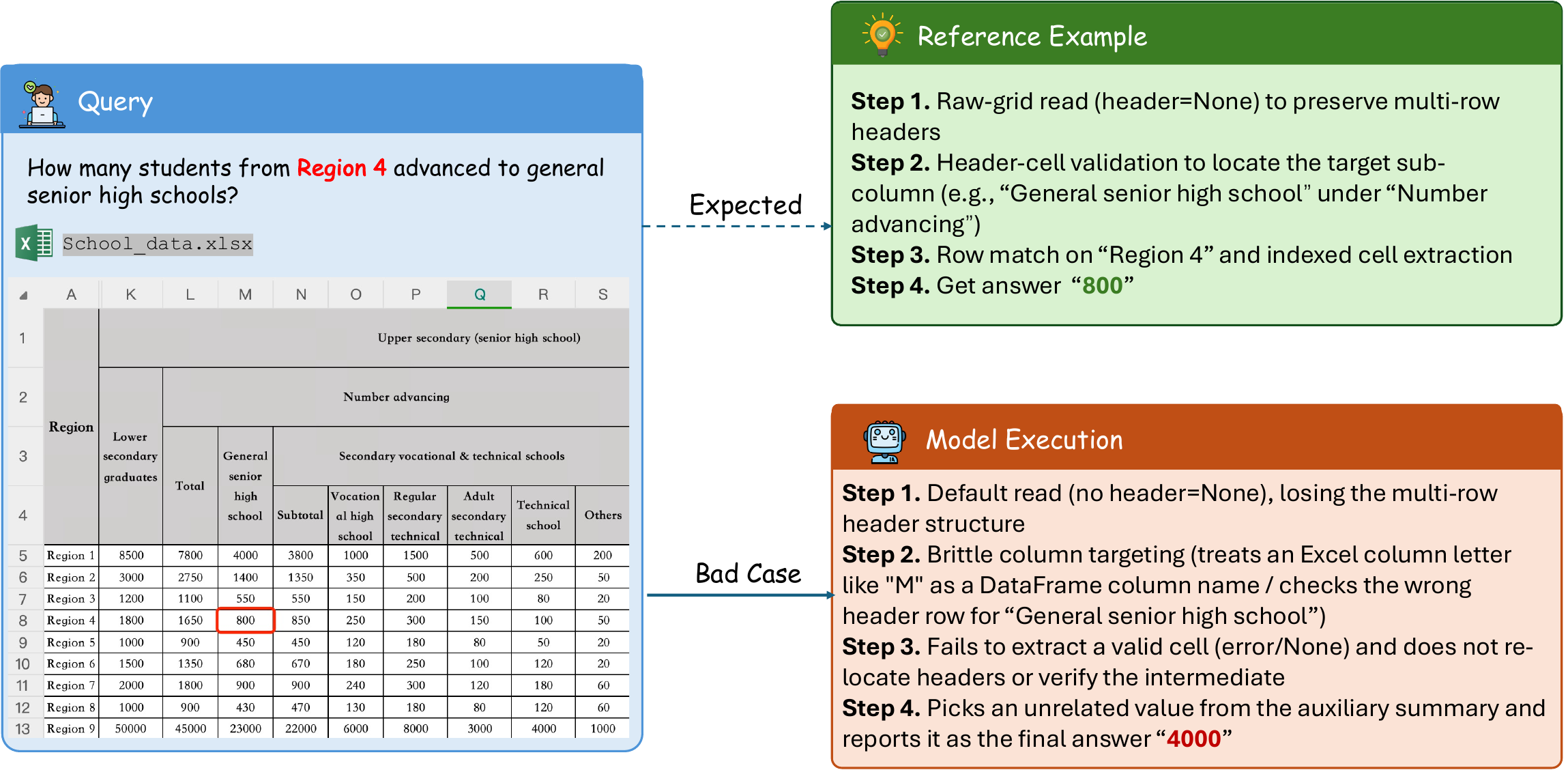}
  \caption{\textbf{Numerical calculation error due to hierarchical layout parsing failure.} 
The figure demonstrates a failure mode where the model is unable to correctly interpret the multi-row header structure of the spreadsheet. 
While the reference approach accurately locates the target data (highlighted in the red box), the model misaligns the table structure and erroneously retrieves a value from an unrelated row.}
\label{fig:numerical_failure}
\end{figure*}

Figure~\ref{fig:numerical_failure} shows a numerical calculation error driven by brittle spreadsheet parsing. 
The task requires extracting a single scalar from an Excel file (e.g., Region~4 $\rightarrow$ General High School). 
The reference pipeline correctly resolves the header hierarchy, selects the target sub-column, filters Region~4, and returns the ground truth (\textbf{800}). 

In contrast, the model assumes Excel column letters map to DataFrame column names and checks an incorrect header row, yielding an invalid intermediate (error/None). 
After multiple failed code attempts to locate the correct cell, it directly picks a value from the auxiliary summary and reports it as the final answer (\textbf{4000}). 
This failure highlights the necessity of semantic-aware parsing (e.g., `header=None` grid analysis) over default DataFrame loading when dealing with multi-row headers.

\end{document}